\documentclass[Journal]{IEEEtran}
\IEEEoverridecommandlockouts
\usepackage{amsmath,amssymb,amsfonts}
\usepackage{algorithmic}
\usepackage{graphicx}
\usepackage{textcomp}
\usepackage{xcolor}
\usepackage[caption=false]{subfig}
\usepackage{multirow}
\usepackage{multicol}
\usepackage{epsfig}
\usepackage{float}
\usepackage{url}
\usepackage{lscape}
\usepackage{adjustbox}
\usepackage{hyperref}
\usepackage{cite}

\def\BibTeX{{\rm B\kern-.05em{\sc i\kern-.025em b}\kern-.08em
    T\kern-.1667em\lower.7ex\hbox{E}\kern-.125emX}}

\makeatletter
\def\ps@IEEEtitlepagestyle{%
	\def\@oddfoot{\mycopyrightnotice}%
	\def\@evenfoot{}%
}
\def\mycopyrightnotice{%
 {\hfill \footnotesize xxxxxx  \copyright xxxxxx\hfill}
}
\makeatother

\begin{document}

\title{Generating Automatically Print/Scan Textures for Morphing Attack Detection Applications
\thanks{This work is supported by the European Union’s Horizon 2020 research and innovation program under grant agreement No 883356 and by the German Federal Ministry of Education and Research and the Hessian Ministry of Higher Education, Research, Science and the Arts within their joint support of the National Research Center for Applied Cybersecurity ATHENE.}
}
\author{
\IEEEauthorblockN{Juan E. Tapia, Maximilian Russo and Christoph Busch}\\
\IEEEauthorblockA{\textit{Hochschule Darmstadt} \\
\textit{{da/sec-Biometrics and Internet Security Research Group}}\\
Darmstadt, Germany \\
juan.tapia-farias@h-da.de, maximilian.russo@h-da.de, christoph.busch@h-da.de}\\
\textbf{****This is a pre-print under revision process*****}


}


\maketitle


\begin{abstract}
Morphing Attack Detection (MAD) is a relevant topic that aims to detect attempts by unauthorised individuals to access a "valid" identity. One of the main scenarios is printing morphed images and submitting the respective print in a passport application process. Today, small datasets are available to train the MAD algorithm because of privacy concerns and the limitations resulting from the effort associated with the printing and scanning of images at large numbers.
In order to improve the detection capabilities and spot such morphing attacks, it will be necessary to have a larger and more realistic dataset representing the passport application scenario with the diversity of devices and the resulting printed scanned or compressed images. Creating training data representing the diversity of attacks is a very demanding task because the training material is developed manually. This paper proposes two different methods based on transfer-transfer for automatically creating digital print/scan face images and using such images in the training of a Morphing Attack Detection algorithm. Our proposed method can reach an Equal Error Rate (EER) of 3.84\% and 1.92\% on the FRGC/FERET database when including our synthetic and texture-transfer print/scan with 600 dpi to handcrafted images, respectively. 
\end{abstract}


\section{Introduction}
\label{sec:introduction}

Synthetic face images have been utilised in many fields, as realistic images which can be created with Generative Adversarial Networks (GANs). Most state-of-the-art approaches are based on Convolutional Neural Networks (CNN), transfer learning and GAN to transfer the domain properties from one image to another domain. These methods usually need pairing images of the same objects to perform a pixel transfer style. 

On the other hand, unpaired images can also be used to transfer the style of one image to another image unrelated to or from a different object. In this context, both the pairing and unpairing methods can generate synthetic images to create and transfer the style from printed/scanned images applied to bona fide and morphed face images. 

A large number of such images can support the training of MAD systems, which is a relevant approach to improve the diversity of the training database, which previously consisted of only digital domain images.

MAD is a relevant topic to detect attempts by unauthorised individuals who want to use a "valid" identity document issued to another individual. In recent years, several such cases have been reported, most of them related to illegal border crossings \cite{Torkar-MADcases-2022}. Morphing can be understood as a technique to seamlessly combine two or more look-alike facial images from a subject and an accomplice. A morphing attack takes place in the enrolment process for an identity document. The threat of morphing attacks is known for border crossing or identification control scenarios. It can be broadly divided into two types: (1) Single Image Morphing Attack Detection (S-MAD) techniques and Differential Morphing Attack Detection (D-MAD) methods \cite{Scherhag-MorphingAttacks-Survey-IEEEAccess-2019,Venkatesh161,TapiaSMAD}.

Many countries issue electronic Machine-Readable Travel documents (eMRTD) passports based on the applicant's printed face photo. Some countries offer online portals for passport renewal, where citizens can upload their own face photo \cite{detect_raghu}. However, in most countries, the passport applicant supplies a printed image to the government authority issuing the identity document. It is subsequently scanned and embedded in the identity document, both on the data page and in the chip of the eMRTD. Accepting a printed face image excludes supervision of the face capture process by design; the applicant supplies a facial image, and its provenance cannot be conclusively verified. Then, to detect a morphed image, which was printed and scanned, it will be necessary to train S-MAD and D-MAD systems on a large number of face image samples. 

Developing a MAD system requires data containing samples of the class bona fide (i.e. a pristine image) and also of the class morph (including a large diversity of morphing algorithms) to be used and possibly shared for new research and development. The availability of real biometric face images, such data is limited in diversity and quantity and raises ethical and legal challenges regarding its use and distribution. This fact motivated some researchers to use synthetic-based data to develop MADs \cite{Naser-privacy}. Today, in state of the art, some databases are available to create synthetic morph images, such as Synthetic Morphing Attack Detection Development dataset (SMDD)\cite{Naser-privacy}, MorGAN \cite{MoprhGAN}, MIP-GAN-I and II \cite{MIPGAN} and others. However, these datasets help partially to reduce the detection rate in print/scan scenarios.

In order to improve the previous limitations of a low number of bona fide and morphed images, we improved and adapted two different methods: 
\begin{itemize}
   
\item An image-to-image pairing/unpairing method based on the Pix2pix and CycleGAN algorithms \cite{pix2pix, cycleGan}, which we involved to generate simulated print/scan face images. GitHub \footnote{\url{https://github.com/jedota/Style-Transfer-PS600}}. 
\item  A second semi-automatic method to isolate the noise/artefact created in the print/scan process by the hardware used for this task and apply it directly to bona fide and morph images. GitHub:\footnote{\url{https://github.com/jedota/Semi-Auto-Transfer-PS600}}.

\end{itemize}

It is essential to highlight that the traditional method of manually creating print/scan images is very time-consuming. 

We describe the scenario as follows: First, a set of bona fide and morphed images is selected. Second, a printed version of digital images is created using high-quality glossy paper. Afterwards, printed images are re-digitised with a desktop scanner. Later, these images are manually checked for quality, occlusion or other artefacts. It is essential to highlight that this process must be repeated image by image for each new scanning device, which complies with the technical regulations defining the passport application process.


In summary, the contributions of this work are as follows.

\begin{itemize}
\item One method based on the transfer-style technique was proposed to create synthetic digital print/scan on 600 dpi versions from digital images.
\item Four different backbones models based on UNet128, UNet256 and ResNet50 (conv-layer 6 and conv-layer 9) were trained from scratch and evaluated to get high-quality face images. 
\item A second semi-automatic texture-transfer method based on computer vision was also proposed to simulate artificial print/scan texture.
\item The output images from both previous methods were evaluated based on Frechet Inception Distance to measure the similarity of the image generated in comparison with the source image.
\item In order to show the relevance of the method, a Leave-One-Out protocol was applied to S-MAD based on the FRGC/FERET databases using a Support Vector Machine (SVM) classifier with four morphing tools to show the utility of our proposed method in terms of high explicitly S-MAD accuracy. In the end, we performed 144 evaluations in total.
\item The image results from both methods, the transfer-style process and texture-transfer, allow us to improve the S-MAD results regarding Equal Error Rate (EER) with a mixture in training of real images plus generated images.
\item All the methods presented in this work are fully reproducible. The GitHub implementation will be available (Upon acceptance).
\end{itemize}
 
The rest of the article is organised as follows: Section~\ref{sec:related} summarises the related work on MAD. The database description is explained in Section~\ref{sec:dataset}. The metrics are explained in Section~\ref{sec:metrics}. The experiment and results are presented in Section~\ref{sec:results}. We conclude the article in Section~\ref{sec:conclusions}.


\section{Related Work}
\label{sec:related}

\subsection{Generative Adversarial Network}
Most of the approaches for image-to-image translation that are recently reported in the literature use transfer techniques based on Deep Learning (DL), where GAN are applied to the input image in order to translate the content to the target domain \cite{Mitkovski}.

Zhu et al.~\cite{cycleGan} proposed an approach for learning to translate an image from a source domain $X$ to a target domain $Y$ in the absence of paired examples. The goal is to learn a mapping $G: X$ $\rightarrow$ $Y$ such that the distribution of images from $G(X)$ is indistinguishable from the distribution $Y$ using an adversarial loss. Qualitative results are presented on several tasks where paired training data does not exist, including collection style transfer, season transfer, and photo enhancement.

Gatys et al.~\cite{NIPS2015} proposed a new parametric texture model to tackle this problem. Instead of describing textures based on a model for the early visual system, they use a CNN – a functional model for the entire ventral stream – as the foundation for this texture model. The feature information is extracted by 16 convolutional and five pooling layers.

Johnson et al.~\cite{Johnson} proposed utilising perceptual loss functions to train feed-forward networks for real-time texture transfer tasks. Li and Wand \cite{Li2016CombiningMR} combined the Markov Random Fields model with deep neural networks, which was later extended to semantic style transfer.

Karras et al.~\cite{stylegan2} developed the StyleGAN3 network \footnote{\url{https://github.com/NVlabs/stylegan3}}. This GAN is an extension of the progressive growing GAN that is an approach for training generator models capable of synthesising huge high-quality images via the incremental expansion of discriminator and generator models from small to large images during the training process. 

Very recently, Markham et al. \cite{markham2023openset}, studied the performance in the open-set datasets of smartphone video clips containing bona fide ID Cards and created print and screen presentation attack samples. He used 4 different transfer style methods for this task to simulate presentation attacks.

\subsection{Morphing Attack Detection}

Regarding MAD oriented to detect print/scan, Raghavendra et al.~\cite{raghu_cvpr} explored the transfer learning approach using the Deep Convolutional Neural Networks (D-CNN) to detect both digital and print-scanned versions of morphed face images. This work explored a feature-level fusion of the first fully connected layer from pre-trained VGG19 and AlexNet networks\cite{alexnet}. The database contains only 352 bona fide and 431 morphed images corresponding to digital and print-scanned versions.

Debiasi et al.~\cite{Debiasi-MorphingDetectionPRNU-IWBF-2018} and
Scherhag et al.~\cite{Scherhag-PRNU-TBIOM-2019} proposed to exploit the image noise patterns by Photo Response Non-Uniformity (PRNU) analysis, where the unique PRNU-pattern is extracted and analysed.

Mitkovski et al. ~\cite{Mitkovski} proposed a method based on a conditional generative adversarial network to generate print and scan images. The goodness of simulation is evaluated with respect to image quality, biometric sample quality and performance, and human assessment.

Ferrara et al.~\cite{Mateo_print} proposed an approach based on Deep Neural Networks for morphing attack detection. In particular, the generation of simulated printed‐scanned images and other data augmentation strategies and pre‐training on large face recognition datasets. The author used the Progressive Morphing Database (PMDB) ~\cite{PSMB_data} for network training. This dataset contains 6,000 morphed images automatically generated starting from 280 subjects selected from a different dataset.

Damer et al.~\cite{pixelwise} proposed a pixel-wise morphing attack detection (PW-MAD) approach where they train a network to classify each pixel of the image rather than only having one label for the whole image. Additionally, they created a new face morphing attack dataset with digital and re-digitised samples, namely the LMA-DRD dataset. However, this dataset presents limited printed and scanned images with only 276 bona fide attacks.

In our previous work \cite{Tapia-EUSIPCO}, we recently proposed a basic version to create synthetic texture using the Pix2pix transfer style only in the FRGC dataset using Random Forest and MobileNetv2. Thus, motivated by the previous results, we will present an extended version that proposes new characteristics, such as two transfer-style methods based on GANs and one semi-automatic handcrafted method based on 50 texture palette colours on FRGC/FERET datasets and its application to S-MAD using features extracted from Intensity, Shape, Frequency and Compression.

\section{Databases}
\label{sec:dataset}
In this work, we employ the FRGCv2\cite{FRGC} and FERET databases \cite{Phillips-FERET-1998} for our experiments. The selection of these databases was motivated because we developed the morphed images in two sets: No post-processing and the print/scan at 600 dpi version, which means printing and scanning images one by one was done with a resolution of 600 dpi. The original images have a size of $360\times480$ pixels. 

The databases were processed twice for different stages. First, synthetic images were created using transfer-style and texture-transfer algorithms. We developed unpairs and pairs of images side by side to represent the original digital images and the manually created printed/scanned version in 600 dpi. For pairing images, both images must have the same subject. Different subjects were used for unpaired images. For both datasets, the size and image alignment are the same. The side-by-size images have a final size of $720\times480$ pixels. 

As a second process, the FRGCv2, FERET database and the output of both proposed method-based transfer style and texture-transfer were used to train a classifier based on SVM with RBF-Kernel to detect bona fide and morph images. With this setup, it is possible to evaluate the influence of our proposed methods. All the images were aligned and cropped by MTCNN library \cite{MTCNN}.

The following algorithms were used to create the morphed images:
\begin{itemize}
    \item FaceFusion is a proprietary morphing algorithm developed for IOS app \footnote{\url{www.wearemoment.com/FaceFusion/}}. This algorithm creates high-quality morph images without visible artefacts.
    \item FaceMorpher is an open-source algorithm to create morph images \footnote{\url{https://github.com/alyssaq/face_morpher}}. This algorithm introduces some artefacts in the background.
    \item OpenCV-Morph, this algorithm is based on the OpenCV implementation \footnote{\url{www.learnopencv.com/face-morph-using-opencv-cpp-python}}. The images contain visible artefacts in the background and some areas of the face.
    \item Face UBO-Morpher \cite{Raja-SOTAMD-DatabaseEvaluationBenchmarking-TIFS-2020}. The University of Bologna developed this algorithm. The resulting images are of high quality without artefacts in the background. 
\end{itemize}

Table \ref{fig:dataset} shows the number of images per dataset and by the morphing tools.
\vspace{-0.3cm}

\begin{table}[H]
\scriptsize
\centering
\caption{Summary for databases and Morphing tools.}

\begin{tabular}{lllll}
\hline
\textbf{Database} & \textbf{Nº Subjects} & \textbf{Bona fide} & \textbf{Morphs} &  \\ \hline
FaceFusion        & 529/984              & 529/984              & 529/964             &             \\ \hline
FaceMorpher       & 529/984             & 529/984             & 529/964             &            \\ \hline
OpenCV-Morph      & 529/984             & 529/984              & 529/964             &             \\ \hline
UBO-Morpher       & 529/984              & 529/984             & 529/964             &            \\ \hline
\end{tabular}
\label{fig:dataset}
\end{table}

\vspace{-0.3cm}
\section{Method}
\label{sec:method}
This section describes the transfer style method implemented based on GANs and also the semi-automatic texture-transfer style as follows:


\subsection{Image Generation using Pix2pix}

\begin{figure}[H]
\centering
\includegraphics[scale=0.40]{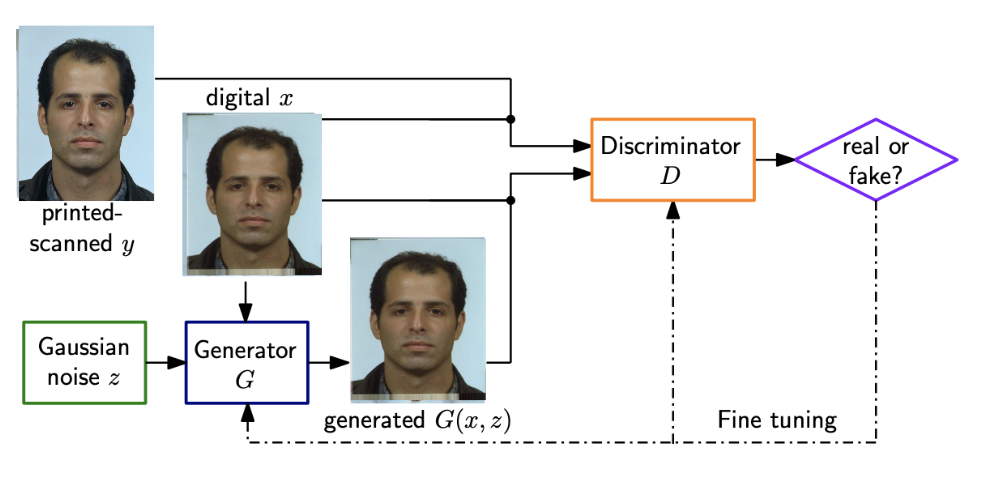}
\caption{Illustration of our pixel-wise print/scan simulation network \cite{Mitkovski}.}
\label{fig:pix2pix}
\end{figure}

One of the methods is based on a transfer style network called pixel-to-pixel (Pix2pix), as illustrated in Figure~\ref{fig:pix2pix}. It takes two face images aligned side-by-side as the input, such as bona fide digital and bona fide from handcrafted print/scan versions from the same subject. These input images are considered paired images. The method delivers as outputs the original images translated to the new domain (print/scan). This is depicted as the two images passing through two identical copies of the CNN block, but in practice, only one copy is stored in memory. It takes the two images, producing the two embeddings. 

The Pix2pix is based directly on the conditional GAN architecture. It is used as input to both networks, concatenating the original input and conditioning images on the channel dimension. In addition to the GAN loss, we used a $L1$ loss to enforce correctness at the low frequencies. A U-Net \cite{ronneberger2015unet} architecture is used as a baseline generator, while a ``PatchGAN'' architecture is used as the discriminator that classifies patches as real or fake and aggregates the results. Even though the use of PatchGAN of $60\times60$ is more efficient in terms of training time, this implementation generated localised noises and artefacts in the faces. Because of that, we modified the original implementation and used a pixel-wise approach. The results show that this new approach can reach a lower FID. See Table~\ref{tab:fid_score}.

\subsection{Image Generation using CycleGANs}

CycleGAN achieves unpaired images, which means side-by-side images from two different subjects, and performs the translations from bona fide to print/scan by using two GANs, enforcing a cycle consistency loss between the generators. That is, given the generator $G$ from $\mathcal{X}$ to $\mathcal{Y}$ and the generator $F$ from $\mathcal{Y}$ to $\mathcal{X}$, the authors add to the GAN losses the following in Equation \eqref{eq:1}:
\begin{multline}
\label{eq:1}
    \mathcal{L}_{\text{cyc}}(G, F) = \mathbb{E}_{\mathbf{x}}[|| F(G(\mathbf{x})) - \mathbf{x} ||_{1}] \\ + \mathbb{E}_{\mathbf{y}}[|| G(F(\mathbf{y})) - \mathbf{y} ||_{1}]
\end{multline}
Although the cycle consistency condition is an effective strategy for approaching the unpaired translation problem, it tends to force $G$ to generate samples that contain all the necessary information to translate back to the input image, which leads to unsatisfactory results if significant visual changes are expected. 
We also modify the loss using L1 distance in the loss function rather than L2, as L1 encourages less blurring.

\begin{figure}[h]
\centering
\includegraphics[scale=0.35]{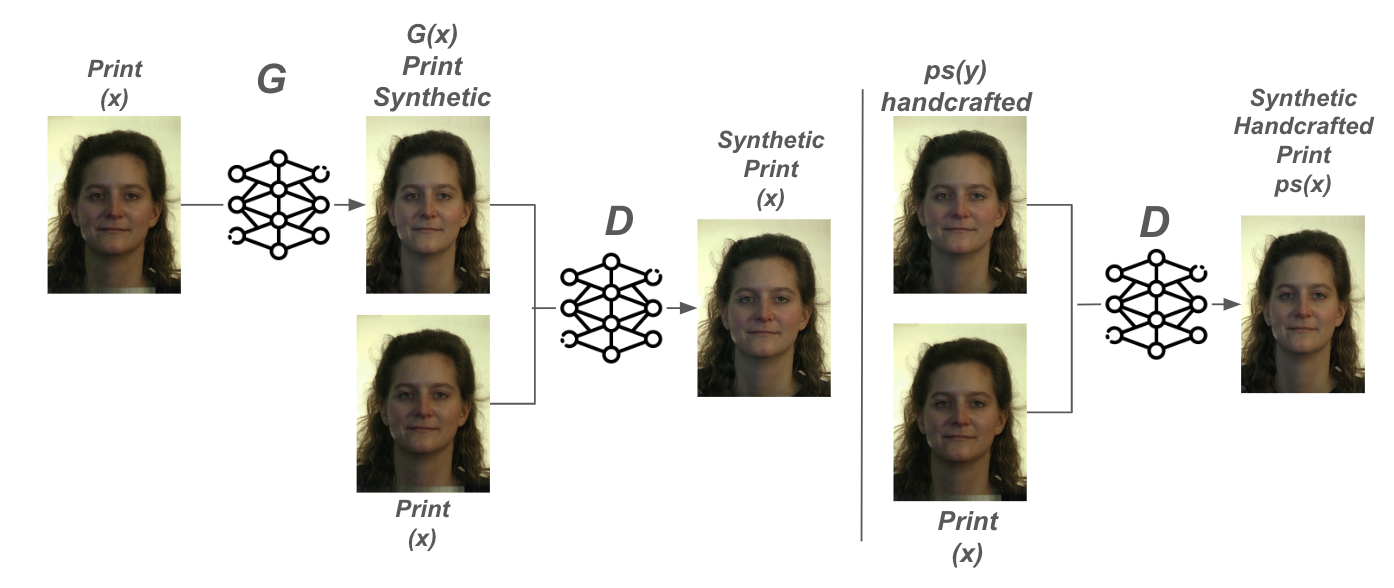}
\caption{Illustration of our CycleGan print/scan simulation network.}
\label{fig:Cycle}
\end{figure}

\begin{figure*}[]
\centering
\includegraphics[scale=0.35]{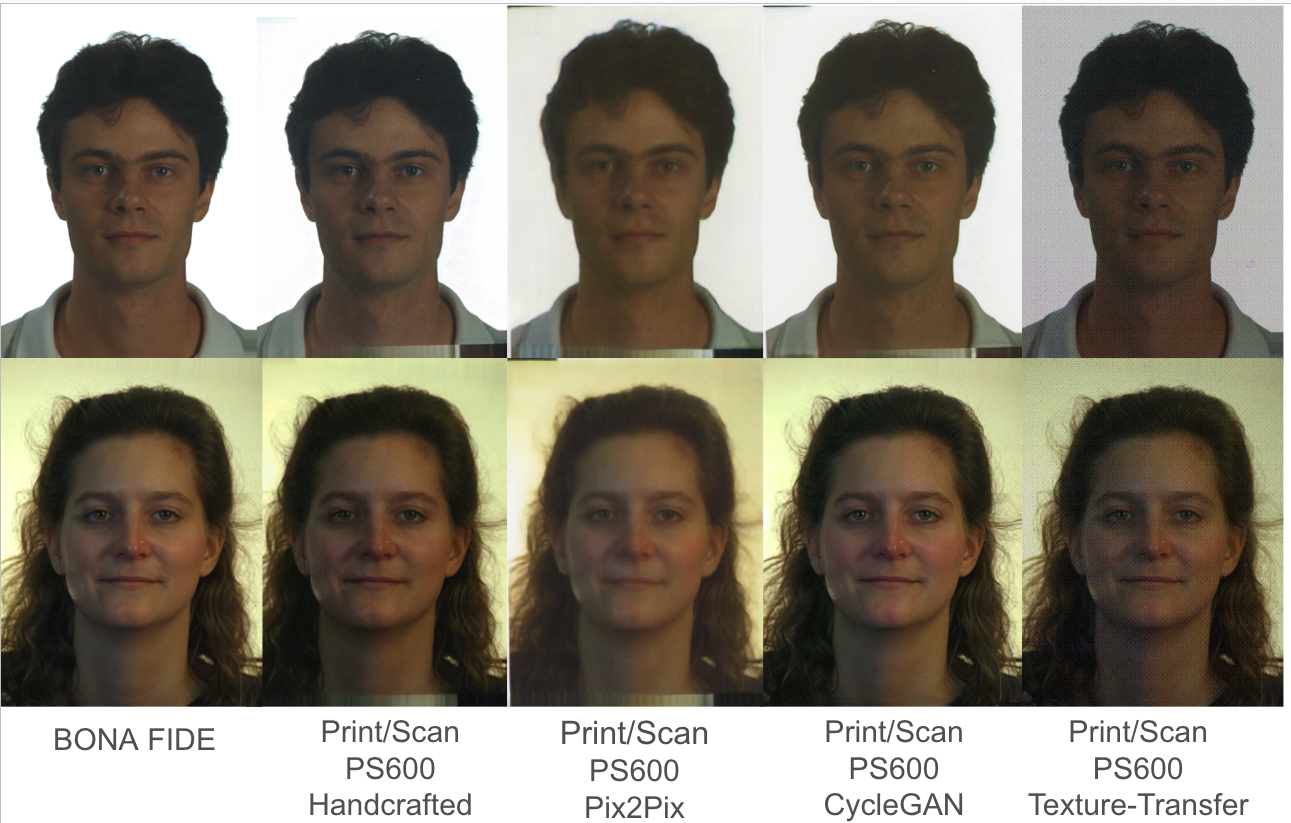}
\caption{Example of all the methods used to generate print/scan images from FRGC dataset.}
\label{fig:GAN_result_image}
\end{figure*}

Figure \ref{fig:Cycle} shows the adapted CycleGAN network to our proposal. Training a conditional GAN to map print/scan images to handcrafted print/scan. The discriminator, $D$, learns to classify between fake (synthesized by the generator) and real {print/scan, handcrafted print/scan} tuples. The generator, $G$, learns to fool the discriminator. Unlike an unconditional GAN, both the generator and discriminator observe the input print/scan image. This network delivers as outputs the original images translated to the new domain (print/scan) as is shown in Figure~\ref{fig:GAN_result_image}.

\vspace{-0.3cm}


\subsection{Image Generation using Handcrafted Texture-Transfer}

Motivated by our previous work \cite{TIFS-Benal}, which applied textures to ID Cards for Presentation Attack Detection focus, we proposed a new second semi-automatic method. This method isolates the texture or artefacts in the capture process caused by sensor noise (hardware), frequency patterns, or moire patterns without any training process. However, at this time, the captured noise/artefact comes from the print and scan hardware instead of images captured by smartphones, which do not represent the real process of requesting a passport. Also, the size of each colour now is equivalent to the size of portrait images used in a passport.

This way, we can quickly transfer this pattern to any new bona fide or morphed image. These kinds of artefacts are inherent to the type of paper, camera sensors, and scanners.

\begin{figure}[H]
\centering
\includegraphics[scale=0.28]{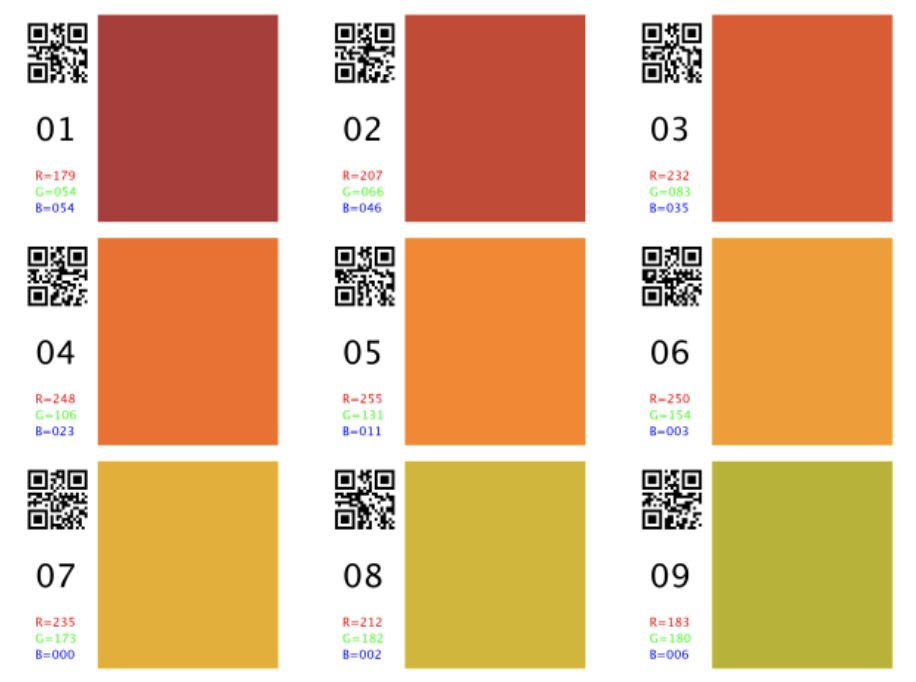}
\caption{Example of 50 texture palettes with the QR for each colour.}
\label{fig:texture_palete}
\end{figure}

In order to do that, we have selected a palette of 50 different colours, as the background colour impacts the texture perception, as shown in Figure~\ref{fig:texture_palete}. 

This palette was captured as raw images, meaning that these colour palettes were manually printed and scanned in 300 and 600 dpi in glossy paper and high-quality using a Brother printer, model L9570.


In order to track a record of the original colour in each image, it placed a QR code next to each colour, describing the original tone. Therefore, we can analyse how the print/scan process shifts this colour and adds the described artefacts in the digital image. We chose solid colours because they would allow us to isolate the texture without any alignment process. This is important since an affine transformation would have added deformation and sampling artefacts.

The next step is to localise the coloured part of the image, remove the background, and isolate the texture that it carries from the printing/scanning process. 


The texture is isolated by simply subtracting the corresponding colour from the image. For this step, we subtracted the colour described by the QR code in the red, green and blue channels.

%


\begin{figure*}[]
\centering
\includegraphics[scale=0.35]{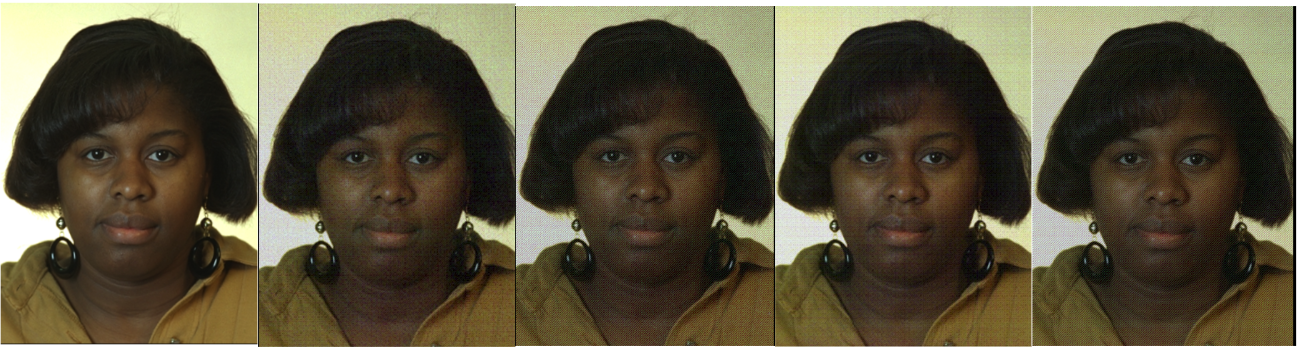}
\caption{Example Texture-transfer technique applied to different sources and resolutions. Left to right: Digital image, Bond paper 300 dpi, Bond paper 600 dpi, Glossy paper 300 dpi and Glossy paper 600 dpi.}
\label{fig:TT-resolution}
\end{figure*}

\section{Evaluation Metrics}
\label{sec:metrics}

\subsection{Sample Quality with Frechet Inception Distance}

One of the difficulties with GAN algorithms, and in particular when applied to face images, is how to assess the quality of the resulting (synthesised) images. Currently, a suite of qualitative and quantitative metrics has been proposed to assess the performance of a GAN model based on the quality and diversity of the generated synthetic images~\cite{SalimansGZCRC16, heusel2017fid}. In this work, we use the Frechet Inception Distance (FID)~\cite{heusel2017fid}. The FID metrics allow us to compare different GAN models with the print/scan image results. The FID score measures the objective quality of the new print/scan synthetic textured images versus the print/scan real images.

Frechet Inception Distance (FID) compares the similarity between two groups of images, A and B. First, to compute the FID, all images from set A and set B have to be processed by an InceptionV3 network, pre-trained on ImageNet~\cite{deng2009imagenet}. Then, the 2,048 feature vector of the Inception-V3-pool3-layer is stored for each image. Finally, the distributions of A and B in the feature space are compared using Equation~\ref{eq:fid}, where $\mu_A$ and $\mu_B$ are the mean values of the distributions A and B, respectively, and $\Sigma_A$ and $\Sigma_B$ are the covariances of the two distributions.
\vspace{-0.3cm}

\begin{equation}\label{eq:fid}
    FID = \| \mu_A - \mu_B \|^2 + Tr \left( \Sigma_A + \Sigma_B -2  ( \Sigma_A \cdot \Sigma_B ) ^{1/2} \right)
\end{equation}


\subsection{Morphing Attack Detection Accuracy}

The ISO/IEC-CD-20059.2 \cite{ISO-IEC-20059} standard presents methodologies for the evaluation of the detection performance of MAD algorithms for biometric systems. The Morphing Attack Classification Error Rate (MACER) metric measures the proportion of Morph attacks for each different morph method incorrectly classified as bona fide presentation. This metric is calculated for each morph, where the worst-case scenario is considered. Equation~\ref{eq:apcer} details how to compute the MACER metric, in which the value of $N_{PAIS}$ corresponds to the number of morph images, where $RES_{i}$ for the $i$th image is $1$ if the algorithm classifies it as a morphed image, or $0$ if it is classified as a bona fide presentation.
\vspace{-0.3cm}

\begin{equation}\label{eq:apcer}
    {MACER_{PAIS}}=1 - (\frac{1}{N_{PAIS}})\sum_{i=1}^{N_{PAIS}}RES_{i}
\end{equation}

Additionally, the Bona Fide Classification Error Rate (BPCER) metric measures the proportion of bona fide presentations mistakenly classified as morphing attack presentations. The BPCER metric is formulated according to equation~\ref{eq:bpcer}, where $N_{BF}$ corresponds to the number of bona fide presentation images, and $RES_{i}$ takes identical values of those of the MACER metric.
\vspace{-0.2cm}

\begin{equation}\label{eq:bpcer}
    BPCER=\frac{\sum_{i=1}^{N_{BF}}RES_{i}}{N_{BF}}
\end{equation}

These metrics effectively measure to what degree the algorithm confuses morphed images with bona fide images and vice versa. The MACER and BPCER metrics depend on a decision threshold.
All the experiments are also reported with a detection error trade-off (DET) curve. In the DET curve, the EER value represents the trade-off when the MACER is equal to the BPCER. Values in this curve are presented as percentages.


\section{Experiments}
\label{sec:results}

This section describes the similarity of the print/scan images generated using the proposed method of creating print/scan face images based on synthetic GANs and print/scan face images using a semi-automatic texture-transfer to images.

\subsection{Experiment 1- Image Generation using GANs}
Portrait pictures from FRGC/FERET were used to train a system to generate new print/scan images. For image generation based on Pix2pix, CycleGAN, and two different convolution neural networks, based on UNet and ResNet50, were explored.

For UNet, the networks UNet128 and UNet256 were trained, and all the images were resized to $128\times128$ and $256\times256$, respectively. The batch size was set to 32 and 200 epochs.

For ResNet, we are fine-tuning the network to improve the results based on convolutional layers block6 and block9. For the input size, the width and height need to be divisible by 4. In our case, we used $512\times512$, and the batch size was set to 16 and 200 epochs.

Table \ref{tab:fid_score} shows the FID scores reached for all the GANs implanted for Pix2pix, CycleGAN and the $\Delta$ for the best methods between baseline FID values and images generated. The first columns show different handcrafted methods and the four morphing tools used. The second column shows the FID value (a lower value for the distance between synthetic print/scan images and real print/scan images is better) between the manual print/scan 600 dpi images and each set of morphing. It is essential to highlight that this value is the goal to reach for our Pix2pix and CycleGAN for automatic print/scan versions. 

Columns three up to five show the FID score reached for Pix2pix-based UNet256, ResNet-6blocks, and ResNet-9blocks. Columns six up to eight show the FID score reached for CycleGAN-based on UNet256, ResNet-6blocks, and ResNet-9blocks. Column 10 reports the FID score of the texture-transfer method. Columns 9 and 11 report the $\Delta$ of the best results, which means lower differences with column 2 were obtained by CycleGAN based on ResNet50-block 9 (column 8) and texture-transfer (column 9). The UNet128 was discarded because it reached very poor results and the highest FID values. An example of images generated by both methods is depicted in Figures \ref{fig:GAN_result_image} and \ref{fig:TT-resolution}.


\begin{table*}[]
\centering
\scriptsize
\caption{Summary results for FID scores applied to FRGC/FERET dataset.}
\label{tab:fid_score}
\begin{tabular}{|cc|ccccccccc|}
\hline
\multicolumn{2}{|c|}{Source} &
  \multicolumn{9}{c|}{FID (↓)} \\ \hline
\multicolumn{1}{|c|}{\begin{tabular}[c]{@{}c@{}}Images\\ Handcrafted\end{tabular}} &
  \begin{tabular}[c]{@{}c@{}}PS600\\ Baseline\\ Handcrafted\end{tabular} &
  \multicolumn{1}{c|}{\begin{tabular}[c]{@{}c@{}}PIX2PIX\\ UNet256\end{tabular}} &
  \multicolumn{1}{c|}{\begin{tabular}[c]{@{}c@{}}PIX2PIX\\ ResNet50\\ Block6\end{tabular}} &
  \multicolumn{1}{c|}{\begin{tabular}[c]{@{}c@{}}PIX2PIX\\ ResNet50\\ Block9\end{tabular}} &
  \multicolumn{1}{c|}{\begin{tabular}[c]{@{}c@{}}CycleGAN\\ UNet256\end{tabular}} &
  \multicolumn{1}{c|}{\begin{tabular}[c]{@{}c@{}}CycleGAN\\ ResNet50\\ Block6\end{tabular}} &
  \multicolumn{1}{c|}{\begin{tabular}[c]{@{}c@{}}CycleGAN\\ ResNet50\\ Block9 (C9)\end{tabular}} &
  \multicolumn{1}{c|}{\begin{tabular}[c]{@{}c@{}} $\Delta$\\ FID baseline\\ FID C9\end{tabular}} &
  \multicolumn{1}{c|}{\begin{tabular}[c]{@{}c@{}}Transfer\\ Texture (TT)\\ Nº7\end{tabular}} &
  \begin{tabular}[c]{@{}c@{}}$\Delta$\\ FID baseline\\ FID TT\end{tabular} \\ \hline
\multicolumn{1}{|c|}{\begin{tabular}[c]{@{}c@{}}Bona fide \\ PS600\end{tabular}} &
  0 &
  \multicolumn{1}{c|}{\textbf{16.98}} &
  \multicolumn{1}{c|}{52.62} &
  \multicolumn{1}{c|}{77.83} &
  \multicolumn{1}{c|}{15.50} &
  \multicolumn{1}{c|}{11.45} &
  \multicolumn{1}{c|}{\textbf{10.87}} &
  \multicolumn{1}{c|}{\textbf{6.11}} &
  \multicolumn{1}{c|}{35.94} &
  \textbf{18.96} \\ \hline
\multicolumn{1}{|c|}{\begin{tabular}[c]{@{}c@{}}FaceMorpher\\ PS600\end{tabular}} &
  63.71 &
  \multicolumn{1}{c|}{\textbf{79.29}} &
  \multicolumn{1}{c|}{100.58} &
  \multicolumn{1}{c|}{117.75} &
  \multicolumn{1}{c|}{69.25} &
  \multicolumn{1}{c|}{64.25} &
  \multicolumn{1}{c|}{\textbf{59.15}} &
  \multicolumn{1}{c|}{20.14} &
  \multicolumn{1}{c|}{83.46} &
  \textbf{4.17} \\ \hline
\multicolumn{1}{|c|}{\begin{tabular}[c]{@{}c@{}}FaceFusion\\ PS600\end{tabular}} &
  114.97 &
  \multicolumn{1}{c|}{\textbf{127.87}} &
  \multicolumn{1}{c|}{186.16} &
  \multicolumn{1}{c|}{166.07} &
  \multicolumn{1}{c|}{93.45} &
  \multicolumn{1}{c|}{85.25} &
  \multicolumn{1}{c|}{\textbf{75.25}} &
  \multicolumn{1}{c|}{52,62} &
  \multicolumn{1}{c|}{50.45} &
  77.74 \\ \hline
\multicolumn{1}{|c|}{\begin{tabular}[c]{@{}c@{}}OpenCV\\ PS600\end{tabular}} &
  28.02 &
  \multicolumn{1}{c|}{\textbf{39.82}} &
  \multicolumn{1}{c|}{75.11} &
  \multicolumn{1}{c|}{107.08} &
  \multicolumn{1}{c|}{35.70} &
  \multicolumn{1}{c|}{32.25} &
  \multicolumn{1}{c|}{\textbf{28.80}} &
  \multicolumn{1}{c|}{11,02} &
  \multicolumn{1}{c|}{45.95} &
  6.13 \\ \hline
\multicolumn{1}{|c|}{\begin{tabular}[c]{@{}c@{}}UBO\\ PS600\end{tabular}} &
  18.55 &
  \multicolumn{1}{c|}{\textbf{20.06}} &
  \multicolumn{1}{c|}{58.11} &
  \multicolumn{1}{c|}{95.84} &
  \multicolumn{1}{c|}{16.95} &
  \multicolumn{1}{c|}{15.65} &
  \multicolumn{1}{c|}{\textbf{11.15}} &
  \multicolumn{1}{c|}{\textbf{8.91}} &
  \multicolumn{1}{c|}{37.36} &
  17.30 \\ \hline
\end{tabular}%
\end{table*}


\subsection{Experiment 2: Image Generation using Handcrafted Features}

\begin{figure}[H]
\centering
\includegraphics[scale=0.23]{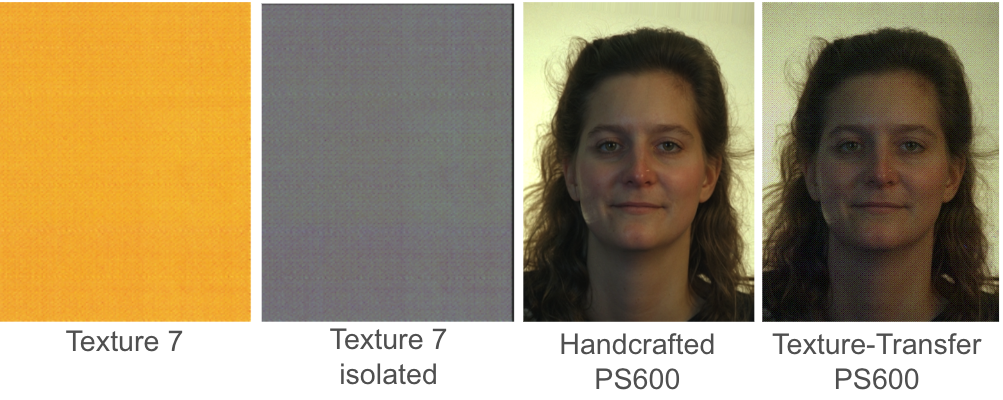}
\caption{Example of application of texture 7 to one digital image.}
\label{fig:ttranfer}
\end{figure}

After each set of the 50 images folder is generated based on the 50-colour palette and applied to all bona fide and morph images from FRGC/FERET datasets, it is necessary to determine which of the 50 textures is the most suitable for our experiment in terms of the FID values. This means we need to look for a colour palette texture that is more similar (lower FID) to the bona fide/morph handcrafted print/scan version images. The best texture-transfer template was the texture number 7 (from 50), with a lower FID of 61.73. Example images for texture 7 are deployed in Figure \ref{fig:ttranfer}.

\begin{figure}[H]
\centering
\includegraphics[scale=0.35]{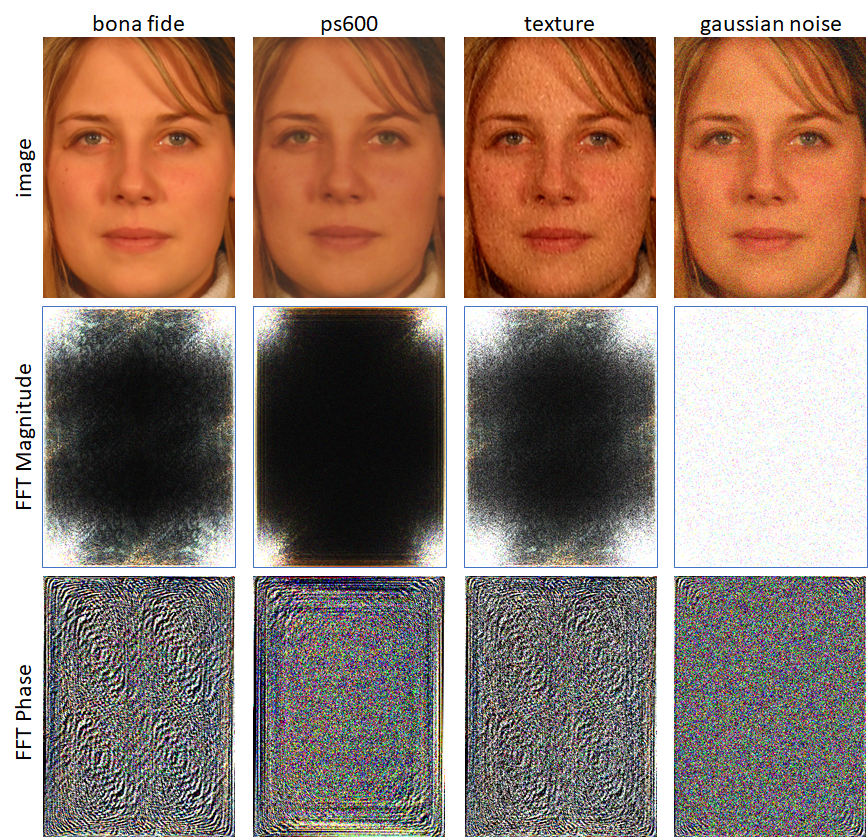}
\caption{Example of FFT image created by synthetic texture and Gaussian Noise.}
\label{fig:fft}
\end{figure}

To demonstrate that the texture produced by our method is different to white or random noise, we have prepared Figure~\ref{fig:fft}, in which the Fast Fourier Transform of bona fide, as well as morph image, is presented. When Gaussian noise is added to the face, green dots appear in the image, which is not present in the other three images. Besides, the spectrum differs from the different morphs in magnitude and phase. 

It is important to highlight that this technique can be applied to different kinds of print papers (bond or glossy) and for different resolutions, such as 300 and 600 dpi.

\section{Application on Morphing Attack Detection}
After generating the different print/scan images with the best models, we applied these new images to the process of training a Single-Morphing Attack Detection. For this task, based on the reduced number of images available for training a SOTA CNN network, we decided to apply our approach to Support Vector Machines based on thirteen features extracted from the Intensity, Shape, Frequency and Compression.

\subsection{Feature extraction methods}

Our intention is to determine which feature would be the most useful and can deliver the most specific information to separate both classes \cite{FF-tapia}. In order to obtain and leverage different features for the bona fide and morphing images, seven different feature extraction methods and several combinations are utilised considering Intensity, Texture, Shape, Frequency and Compression features filter as follows: RAW images (Intensity levels), Discrete Fourier Transform (DFT)~\cite{tan2013digital}, Steganalysis Rich Model (SRM)~\cite{zhou2018learning}, Error Level Analysis (ELA)~\cite{krawetz2007picture}, Single Value Decomposition (SVD), Local Binary Patterns (LBP)\cite{OjalaLPB}, Binary Statistical Image Feature (BISF). These methods are used separately as input for the SVM Classifier and tested against each other.

These feature extraction methods are applied to the original $640\times480$ resolution image, which is then resized and face cropped to the input shape of the network.  This specific order of preprocessing contributes to a better separation of the classes, whereas resizing the image and extracting the features resulted in worse classification performance in all tests.

\subsubsection{Intensity}

For raw data, the intensity of the values in grayscale was used and normalised between 0 and 1. 

\subsubsection{Uniform Local Binary Pattern}

The histogram of uniform LBP and BSIF were used for texture. LBP is a grey-scale texture operator which characterises the spatial structure of the local image texture. Given a central pixel in the image, a binary pattern number is computed by comparing its value with those of its neighbours. 

\begin{equation}
LBP_{P,R}(x,y)=\bigcup_{(x',y')\in N(x,y)}h(I(x,y),I(x',y'))\label{eq:LBP-1}
\end{equation}

where $N(x,y)$ is vicinity around $(x,y)$, $\cup$  is the concatenation operator, $P$ is number of neighbours and $R$ is the radius of the neighbourhood.

\subsubsection {The Binary Statistical Image Feature} was also explored as a texture method. BSIF is a local descriptor designed by binarising the responses to linear filters \cite{BSIF-ICB}. The filters learn from 13 natural images. The code value of pixels is considered a local descriptor of the image intensity pattern in the pixels’ surroundings. The value of each element (i.e bit) in the binary code string is computed by binarising the response of a linear filter with a zero threshold. Each bit is associated with a different filter, and the length of the bit string determines the number of filters used. A grid search from the 60 filters available in BSIF implementation was explored. The filter $5\times5$ and 9 bits obtained the best results estimated from the baseline approach. The resulting BSIF images were used as input for the classifiers.

\subsubsection{Inverse Histogram Oriented Gradient}

For the purpose of describing the shape, the inverse Histogram of oriented gradients \cite{iHOG} was used. The distribution directions of gradients are used as features. Gradients, $x$, and $y$ derivatives of an image are helpful because the magnitude of gradients is large around edges and corners (regions of abrupt intensity changes). We know edges and corners contain more information about object shapes than flat regions. We used the HOG visualisation proposed by Vondrik et al. \cite{iHOG, TapiaSMAD} to select the best parameters to visualise the artefacts in morphed images. This implementation used $10\times12$ blocks and $3\times3$ filter sizes.

\subsubsection{Steganalysis Rich Model}

\begin{figure}[H]
\centering
\includegraphics[scale=0.14]{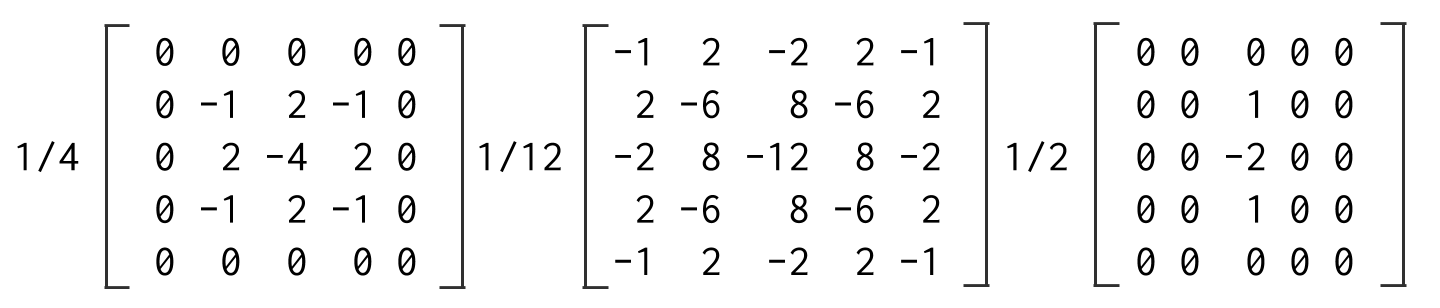}
\caption{SRM filter kernels.}
\label{fig:srmkernel}
\end{figure}

SRM filters yield noise features from neighbouring pixels, which can be applied to detect discrepancies between real and tampered images. The input and output are 3-channel images. As used by Zhou et al.~\cite{zhou2018learning}, the kernels shown in Figure~\ref{fig:srmkernel} are applied to the images, which are then directly used as the input for training the networks.

\subsubsection{Error Level Analysis}
ELA~\cite{krawetz2007picture,luo2010jpeg} is a forensic method to identify portions of an image saved in a JPEG format with a different level of compression. ELA is based on characteristics of image formats that are based on a lossy image compression technique that could be used to determine if an image has been digitally modified. 

JPEG is a method of lossy compression for digital images. The compression level is chosen as a trade-off between image size and quality. A JPEG compression scale is usually 70\%. The compression of data discards or loses information.
The JPEG algorithm works on image grids compressed independently, with a size of $8\times8$ pixels. The $8\times8$ dimension was chosen using a grid search. Meanwhile, any matrices of size less than $8\times8$ do not have enough information, resulting in poor-quality compressed images.

ELA highlights differences in the JPEG compression rate. Regions with uniform colourings, like solid blue or white pixels, will likely have a lower ELA result (darker colour) than high-contrast edges. Highlighted regions can potentially be tampered regions in the image that suffered a second JPEG compression after the user saves the tampered image.

\subsubsection{Discrete Fourier transform}

The discrete Fourier transform (DFT) decomposes a discrete time-domain signal into its frequency components. The purpose of the DFT is to transform the image into its frequency domain representation. The intuition behind this is that differences between the frequencies of multiple face capture devices were used to generate the parent images.

\begin{equation}\label{eq:dft}
   y_{k}=\sum_{n=0}^{N-1}x_{n}e^{-2\pi j\frac{kn}{N}}
\end{equation}

For training purposes, only the magnitude (real) and not the phase (complex) is used. The magnitude image is then transformed from a linear scale to a logarithmic scale to compress the range of values. Furthermore, the quadrants of the matrix are shifted so that zero-value frequencies are placed at the centre of the image. 


\section{Evaluations and Results}
Due to the reduced size of the datasets available for S-MAD evaluation and to avoid overfitting with CNN methods, it was considered an SVM Machine-learning classifier based on feature extractions in a Leave-One-Out(LOO) protocol with four morphing methods. Three evaluations were defined as follows:

\subsection{Evaluation 1- Baseline}

This evaluation considers the baseline because it evaluates the LOO protocol from bona fide images created manually for PS600 versus the Morphed images from FaceFusion, FaceMorpher, OpenCV, and UBO Morpher for FRGC/FERET. 

A LOO protocol was applied to train the morphing attack detection system, which means in the first round, FaceMorpher was used to compute the morphing images used for the test, and training was carried out with OpenCV-Morpher, FaceFusion and UBO-Morpher. Then, in the second round, the OpenCV Morpher was used for testing, and training was performed using morphed images created with FaceFusion, FaceMorpher, UBO-Morpher, and so on. All datasets allow subject-disjoint results to be computed; no subject has an image in both the training and the testing subset. 

In order to train our S-MAD classifier, the FRGC/FERET databases were partitioned to have 70\% training and 30\% validation data. 
Different kinds of features were extracted from faces based on Uniform Local Binary Patterns (uLBP) for all experiments. The histogram of uLBP was used for texture. For the uLBP, all radii values from uLBP81 to uLBP88 were explored (LBP-All). The image's horizontal (uLBP\_HOR) concatenation divided into eight patches was also explored. 

After feature extraction, we fused that information at the feature level by concatenating the feature vectors from different sources into a single feature vector that becomes the input to the classifier. All features were extracted after applying our proposed texture-transfer method. 

Figure \ref{fig:det2024}-Top depicts the DET curves with EER for the baseline results. The results for intensity features reached, on average, an EER of 27.59\%. Texture-based LBP has been reached, on average, an EER of 23.11\%. Frequency based on DCT reached, on average, an EER of 41.37\% for the UBO morph tool.

\subsection{Evaluation 2- Baseline plus Synthetic}

This evaluation considers the baseline images, which means printing/scanning a 600 dpi version of the FRGC/FERET dataset plus the synthetic version created by the best model built from the pix2pix and CycleGAN approach.  The best results were reached with CycleGAN with ResNet9-layers, with an FID of 10.87 for bona fide and 11.75 for morph images. Then, the same LOO protocol was applied in order to analyse and compare the generalisation capabilities of the baseline plus these new synthetic images. The LOO protocol is applied by morph methods considering FaceFusion, FaceMorpher, OpenCV, and UBO Morpher. A detailed summary of results is present in Table \ref{tab:ML-summary} together with Experiment 3 for a direct comparison.

\subsection{Evaluation 3- Baseline plus Texture-Transfer}
This evaluation considers the baseline images, which means printing/scanning a 600 dpi version of the FRGC/FERET dataset plus the semi-automatic texture-transfer technique created by the best filter selected based on the FID score.  The best results were reached with filter number 7, which represents an FID of 35.94 for bona fide and  37.96 for morphing images. 
Then, the same LOO protocol was applied in order to analyse and compare the generalisation capabilities of the baseline plus these new texture-transfer images. The LOO protocol is applied by morph methods considering FaceFusion, FaceMorpher, OpenCV, and UBO Morpher. The summary results on details are present in Table \ref{tab:ML-summary} together with Experiment 2 for a direct comparison. In the end, we performed 144 evaluations in total.
\begin{table*}[]
\centering
\caption{Summary results for all the features extracted for Experiment 2 and Experiment 3. All the results represent the EER (\%).}
\label{tab:ML-summary}
\resizebox{\textwidth}{!}{%
\begin{tabular}{|c|ccccccccccccc|}
\hline
\textbf{\begin{tabular}[c]{@{}c@{}}ONE OUT: \\ FaceFusion\end{tabular}} &
  \multicolumn{13}{c|}{\begin{tabular}[c]{@{}c@{}}Algorithms (EER) -  Exp2 (Syn) / Exp3 (Text)\end{tabular}} \\ \hline
\begin{tabular}[c]{@{}c@{}}Morphing\\ Test\end{tabular} &
  \multicolumn{1}{c|}{RGB} &
  \multicolumn{1}{c|}{ELA} &
  \multicolumn{1}{c|}{SRM} &
  \multicolumn{1}{c|}{DCT2} &
  \multicolumn{1}{c|}{DFT} &
  \multicolumn{1}{c|}{\begin{tabular}[c]{@{}c@{}}LBP\\ 8,1\end{tabular}} &
  \multicolumn{1}{c|}{\begin{tabular}[c]{@{}c@{}}Fusion\\ LBP\end{tabular}} &
  \multicolumn{1}{c|}{HOG} &
  \multicolumn{1}{c|}{SVD} &
  \multicolumn{1}{c|}{HLBP} &
  \multicolumn{1}{c|}{\begin{tabular}[c]{@{}c@{}}BSIF\\ IM\end{tabular}} &
  \multicolumn{1}{c|}{\begin{tabular}[c]{@{}c@{}}BSIF\\ Hist N\end{tabular}} &
  \begin{tabular}[c]{@{}c@{}}\textbf{Average} \\ by \\ \textbf{Morph} \\ Method\end{tabular} \\ \hline
FaceMorpher &
  \multicolumn{1}{c|}{8.26/2.44} &
  \multicolumn{1}{c|}{7.97/2.17} &
  \multicolumn{1}{c|}{0.76/49.64} &
  \multicolumn{1}{c|}{4.74/2.59} &
  \multicolumn{1}{c|}{7.24/0.55} &
  \multicolumn{1}{c|}{37.04/3.52} &
  \multicolumn{1}{c|}{26.64/2.55} &
  \multicolumn{1}{c|}{21.83/4.51} &
  \multicolumn{1}{c|}{7.81/5.47} &
  \multicolumn{1}{c|}{12.85/1.47} &
  \multicolumn{1}{c|}{15.28/2.42} &
  \multicolumn{1}{c|}{20.45/2.29} &
  14.23/6.63 \\ \hline
OpenCV &
  \multicolumn{1}{c|}{7.912.48} &
  \multicolumn{1}{c|}{7.65/2.22} &
  \multicolumn{1}{c|}{0.92/49.62} &
  \multicolumn{1}{c|}{5.60/3.07} &
  \multicolumn{1}{c|}{5.11/0.10} &
  \multicolumn{1}{c|}{36.22/1.09} &
  \multicolumn{1}{c|}{27.74/1.79} &
  \multicolumn{1}{c|}{19.83/2.86} &
  \multicolumn{1}{c|}{7.71/5.18} &
  \multicolumn{1}{c|}{11/901.20} &
  \multicolumn{1}{c|}{6.34/1.33} &
  \multicolumn{1}{c|}{10.30/2.16} &
  \textbf{12.26/6.09} \\ \hline
UBO &
  \multicolumn{1}{c|}{12.08/3.76} &
  \multicolumn{1}{c|}{11.59/3.50} &
  \multicolumn{1}{c|}{0.82/49.62} &
  \multicolumn{1}{c|}{6.17/3.16} &
  \multicolumn{1}{c|}{11.22/0.98} &
  \multicolumn{1}{c|}{35.64/1.64} &
  \multicolumn{1}{c|}{33.02/4.28} &
  \multicolumn{1}{c|}{20.80/2.90} &
  \multicolumn{1}{c|}{12.59/9.26} &
  \multicolumn{1}{c|}{14.91/2.39} &
  \multicolumn{1}{c|}{13.75/1.91} &
  \multicolumn{1}{c|}{18.41/2.19} &
  15.91/7.13 \\ \hline
\begin{tabular}[c]{@{}c@{}}\textbf{Average By} \\ \textbf{Feature} \end{tabular} &
  \multicolumn{1}{c|}{9.42/2.89} &
  \multicolumn{1}{c|}{9.07/2.63} &
  \multicolumn{1}{c|}{0.83/49.6} &
  \multicolumn{1}{c|}{\textbf{5.50/2.94}} &
  \multicolumn{1}{c|}{7.85/0.54} &
  \multicolumn{1}{c|}{36.30/2.08} &
  \multicolumn{1}{c|}{29.13/2.87} &
  \multicolumn{1}{c|}{20.82/3.42} &
  \multicolumn{1}{c|}{9.37/6.63} &
  \multicolumn{1}{c|}{13.22/1.68} &
  \multicolumn{1}{c|}{11.79/1.89} &
  \multicolumn{1}{c|}{16.39/2.21} &
  14.14/6.61 \\ \hline
\textbf{\begin{tabular}[c]{@{}c@{}}ONE OUT: \\ FaceMorpher\end{tabular}} &
  \multicolumn{13}{c|}{\begin{tabular}[c]{@{}c@{}}Algorithms (EER) - Exp2 (Syn) / Exp3 (Text)\end{tabular}} \\ \hline
\begin{tabular}[c]{@{}c@{}}Morphing\\ Test\end{tabular} &
  \multicolumn{1}{c|}{RGB} &
  \multicolumn{1}{c|}{ELA} &
  \multicolumn{1}{c|}{SRM} &
  \multicolumn{1}{c|}{DCT2} &
  \multicolumn{1}{c|}{DFT} &
  \multicolumn{1}{c|}{\begin{tabular}[c]{@{}c@{}}LBP\\ 8,1\end{tabular}} &
  \multicolumn{1}{c|}{\begin{tabular}[c]{@{}c@{}}Fusion \\ LBP\end{tabular}} &
  \multicolumn{1}{c|}{HOG} &
  \multicolumn{1}{c|}{SVD} &
  \multicolumn{1}{c|}{HLBP} &
  \multicolumn{1}{c|}{\begin{tabular}[c]{@{}c@{}}BSIF\\ IM\end{tabular}} &
  \multicolumn{1}{c|}{\begin{tabular}[c]{@{}c@{}}BSIF \\ Hist N\end{tabular}} &
  \begin{tabular}[c]{@{}c@{}}\textbf{Average}\\ by \\ \textbf{Morph} \\ Method\end{tabular} \\ \hline
FaceFusion &
  \multicolumn{1}{c|}{13.61/5.60} &
  \multicolumn{1}{c|}{13.03/4.97} &
  \multicolumn{1}{c|}{0.86/2.49} &
  \multicolumn{1}{c|}{5.0/1.33} &
  \multicolumn{1}{c|}{50.0/50.0} &
  \multicolumn{1}{c|}{18.13/1.91} &
  \multicolumn{1}{c|}{24.62/4.77} &
  \multicolumn{1}{c|}{17.20/5.85} &
  \multicolumn{1}{c|}{14.77/9.78} &
  \multicolumn{1}{c|}{15.58/6.58} &
  \multicolumn{1}{c|}{11.36/8.22} &
  \multicolumn{1}{c|}{26.05/7.02} &
  17.51/9.03 \\ \hline
OpenCV &
  \multicolumn{1}{c|}{4.63/1.82} &
  \multicolumn{1}{c|}{4.57/1.63} &
  \multicolumn{1}{c|}{0.95/2.41} &
  \multicolumn{1}{c|}{3.13/0.90} &
  \multicolumn{1}{c|}{50.0/50.0} &
  \multicolumn{1}{c|}{14.58/0.63} &
  \multicolumn{1}{c|}{17.69/2.51} &
  \multicolumn{1}{c|}{11.08/3.41} &
  \multicolumn{1}{c|}{4.06/2.62} &
  \multicolumn{1}{c|}{7.29/1.84} &
  \multicolumn{1}{c|}{7.42/2.06} &
  \multicolumn{1}{c|}{21.20/2.75} &
  \textbf{12.21/6.04} \\ \hline
UBO &
  \multicolumn{1}{c|}{14.69/6.35} &
  \multicolumn{1}{c|}{14.39/5.55} &
  \multicolumn{1}{c|}{1.07/2.38} &
  \multicolumn{1}{c|}{7.071.56} &
  \multicolumn{1}{c|}{50.0/50.0} &
  \multicolumn{1}{c|}{16.75/1.99} &
  \multicolumn{1}{c|}{26.55/5.44} &
  \multicolumn{1}{c|}{16.60/5.18} &
  \multicolumn{1}{c|}{15.77/10.16} &
  \multicolumn{1}{c|}{14.96/5.37} &
  \multicolumn{1}{c|}{9.97/6.92} &
  \multicolumn{1}{c|}{27.41/5.49} &
  17.93/8.74 \\ \hline
\begin{tabular}[c]{@{}c@{}}\textbf{Average by} \\ \textbf{Feature} \end{tabular} &
  \multicolumn{1}{c|}{10.98/4.59} &
  \multicolumn{1}{c|}{10.66/3.51} &
  \multicolumn{1}{c|}{\textbf{0.96/2.41}} &
  \multicolumn{1}{c|}{\textbf{5.06/1.26}} &
  \multicolumn{1}{c|}{50.00/50.0} &
  \multicolumn{1}{c|}{16.49/1.51} &
  \multicolumn{1}{c|}{22.95/4.24} &
  \multicolumn{1}{c|}{14.96/4.81} &
  \multicolumn{1}{c|}{11.53/7.52} &
  \multicolumn{1}{c|}{12.61/4.59} &
  \multicolumn{1}{c|}{9.58/5.73} &
  \multicolumn{1}{c|}{24.89/5.09} &
  {15.89/7.94} \\ \hline
\textbf{\begin{tabular}[c]{@{}c@{}}ONE OUT: \\ OpenCV\end{tabular}} &
  \multicolumn{13}{c|}{\begin{tabular}[c]{@{}c@{}}Algorithms (EER) - Exp2 (Syn) / Exp3 (Text)\end{tabular}} \\ \hline
\begin{tabular}[c]{@{}c@{}}Morphing\\ Test\end{tabular} &
  \multicolumn{1}{c|}{RGB} &
  \multicolumn{1}{c|}{ELA} &
  \multicolumn{1}{c|}{SRM} &
  \multicolumn{1}{c|}{DCT2} &
  \multicolumn{1}{c|}{DFT} &
  \multicolumn{1}{c|}{\begin{tabular}[c]{@{}c@{}}LBP\\ 8,1\end{tabular}} &
  \multicolumn{1}{c|}{\begin{tabular}[c]{@{}c@{}}Fusion\\ LBP\end{tabular}} &
  \multicolumn{1}{c|}{HOG} &
  \multicolumn{1}{c|}{SVD} &
  \multicolumn{1}{c|}{HLBP} &
  \multicolumn{1}{c|}{\begin{tabular}[c]{@{}c@{}}BSIF\\ IM\end{tabular}} &
  \multicolumn{1}{c|}{\begin{tabular}[c]{@{}c@{}}BSIF\\ Hist N\end{tabular}} &
  \begin{tabular}[c]{@{}c@{}}\textbf{Average}\\ by \\ \textbf{Morph} \\ Method\end{tabular} \\ \hline
FaceFusion &
  \multicolumn{1}{c|}{13.65/4.93} &
  \multicolumn{1}{c|}{13.47/4.20} &
  \multicolumn{1}{c|}{1.02/0.0} &
  \multicolumn{1}{c|}{7.54/1.42} &
  \multicolumn{1}{c|}{26.18/1.81} &
  \multicolumn{1}{c|}{38.18/0.0} &
  \multicolumn{1}{c|}{38.28/3.43} &
  \multicolumn{1}{c|}{18.80/3.89} &
  \multicolumn{1}{c|}{15.18/9.50} &
  \multicolumn{1}{c|}{19.84/2.56} &
  \multicolumn{1}{c|}{11/364.68} &
  \multicolumn{1}{c|}{23.99/2.11} &
  18.95/3.12 \\ \hline
FaceMorpher &
  \multicolumn{1}{c|}{5.33/1.94} &
  \multicolumn{1}{c|}{5.33/1.55} &
  \multicolumn{1}{c|}{0.88/0.0} &
  \multicolumn{1}{c|}{3.18/0.85} &
  \multicolumn{1}{c|}{16.66/0.0} &
  \multicolumn{1}{c|}{35.67/0.10} &
  \multicolumn{1}{c|}{29.84/2.22} &
  \multicolumn{1}{c|}{15.75/3.57} &
  \multicolumn{1}{c|}{4.28/2.61} &
  \multicolumn{1}{c|}{10.44/0.41} &
  \multicolumn{1}{c|}{7.74/2.19} &
  \multicolumn{1}{c|}{13.040.49} &
  \textbf{12.34/1.32} \\ \hline
UBO &
  \multicolumn{1}{c|}{14.50/5.50} &
  \multicolumn{1}{c|}{14.45/4.61} &
  \multicolumn{1}{c|}{1.14/0.0} &
  \multicolumn{1}{c|}{8.78/1.58} &
  \multicolumn{1}{c|}{24.93/1.65} &
  \multicolumn{1}{c|}{38.06/0.22} &
  \multicolumn{1}{c|}{341.74/4.05} &
  \multicolumn{1}{c|}{18.10/4.25} &
  \multicolumn{1}{c|}{15.52/10.24} &
  \multicolumn{1}{c|}{19.35/1.16} &
  \multicolumn{1}{c|}{9.97/3.75} &
  \multicolumn{1}{c|}{22.22/1.36} &
  18.43/3.19 \\ \hline
\begin{tabular}[c]{@{}c@{}}\textbf{Average by} \\ \textbf{Feature} \end{tabular} &
  \multicolumn{1}{c|}{11.16/4.12} &
  \multicolumn{1}{c|}{11.08/3.45} &
  \multicolumn{1}{c|}{\textbf{1.01/0.0}} &
  \multicolumn{1}{c|}{\textbf{6.5/1.28}} &
  \multicolumn{1}{c|}{22.59/1.15} &
  \multicolumn{1}{c|}{37.30/0.11} &
  \multicolumn{1}{c|}{34.09/3.23} &
  \multicolumn{1}{c|}{17.55/3.90} &
  \multicolumn{1}{c|}{11.66/7.45} &
  \multicolumn{1}{c|}{16.54/1.38} &
  \multicolumn{1}{c|}{9.69/3.19} &
  \multicolumn{1}{c|}{19.75/1.32} &
  {16.57/2.54} \\ \hline
\textbf{\begin{tabular}[c]{@{}c@{}}ONE OUT: \\ Morph UBO\end{tabular}} &
  \multicolumn{13}{c|}{\begin{tabular}[c]{@{}c@{}}Algorithms (EER) - Exp2 (Syn) / Exp3 (Text)\end{tabular}} \\ \hline
\begin{tabular}[c]{@{}c@{}}Morphing\\ Test\end{tabular} &
  \multicolumn{1}{c|}{RGB} &
  \multicolumn{1}{c|}{ELA} &
  \multicolumn{1}{c|}{SRM} &
  \multicolumn{1}{c|}{DCT2} &
  \multicolumn{1}{c|}{DFT} &
  \multicolumn{1}{c|}{\begin{tabular}[c]{@{}c@{}}LBP\\ 8,1\end{tabular}} &
  \multicolumn{1}{c|}{\begin{tabular}[c]{@{}c@{}}Fusion\\ LBP\end{tabular}} &
  \multicolumn{1}{c|}{HOG} &
  \multicolumn{1}{c|}{SVD} &
  \multicolumn{1}{c|}{HLBP} &
  \multicolumn{1}{c|}{\begin{tabular}[c]{@{}c@{}}BSIF\\ IM\end{tabular}} &
  \multicolumn{1}{c|}{\begin{tabular}[c]{@{}c@{}}BSIF\\ Hist N\end{tabular}} &
  \begin{tabular}[c]{@{}c@{}}Average\\ by \\ Morph \\ Method\end{tabular} \\ \hline
FaceFsuion &
  \multicolumn{1}{c|}{10.84/5.01} &
  \multicolumn{1}{c|}{10.54/4.91} &
  \multicolumn{1}{c|}{0.99/0.82} &
  \multicolumn{1}{c|}{4.31/2.26} &
  \multicolumn{1}{c|}{15.88/1.65} &
  \multicolumn{1}{c|}{28.53/1.12} &
  \multicolumn{1}{c|}{27.35/2.60} &
  \multicolumn{1}{c|}{18.35/3.06} &
  \multicolumn{1}{c|}{10.62/8.85} &
  \multicolumn{1}{c|}{12.48/1.12} &
  \multicolumn{1}{c|}{8.18/2.54} &
  \multicolumn{1}{c|}{14.16/1.84} &
  13.51/2.98 \\ \hline
FaceMorpher &
  \multicolumn{1}{c|}{8.65/3.64} &
  \multicolumn{1}{c|}{8.45/3.40} &
  \multicolumn{1}{c|}{0.96/0.85} &
  \multicolumn{1}{c|}{3.60/1.76} &
  \multicolumn{1}{c|}{12.40/1.06} &
  \multicolumn{1}{c|}{30.16/2.82} &
  \multicolumn{1}{c|}{24.31/2.11} &
  \multicolumn{1}{c|}{19.05/4.94} &
  \multicolumn{1}{c|}{7.63/6.13} &
  \multicolumn{1}{c|}{11.33/0.96} &
  \multicolumn{1}{c|}{8.0/2.20} &
  \multicolumn{1}{c|}{13.26/1.59} &
  12.32/2.62 \\ \hline
OpenCV &
  \multicolumn{1}{c|}{7.91/3.52} &
  \multicolumn{1}{c|}{7.84/3.36} &
  \multicolumn{1}{c|}{1.06/0.79} &
  \multicolumn{1}{c|}{3.62/1.73} &
  \multicolumn{1}{c|}{7.79/0.38} &
  \multicolumn{1}{c|}{30.35/1.05} &
  \multicolumn{1}{c|}{27.31/1.93} &
  \multicolumn{1}{c|}{16.20/2.90} &
  \multicolumn{1}{c|}{7.37/5.54} &
  \multicolumn{1}{c|}{10.57/0.73} &
  \multicolumn{1}{c|}{5.16/1.36} &
  \multicolumn{1}{c|}{13.61/1.67} &
  \textbf{11.56/2.08} \\ \hline
\begin{tabular}[c]{@{}c@{}}\textbf{Average by}\\ \textbf{Feature} \end{tabular} &
  \multicolumn{1}{c|}{9.13/4.06} &
  \multicolumn{1}{c|}{8.94/3.9} &
  \multicolumn{1}{c|}{\textbf{1.00/0.82}} &
  \multicolumn{1}{c|}{\textbf{3.84/1.92}} &
  \multicolumn{1}{c|}{12.02/1.03} &
  \multicolumn{1}{c|}{29.68/1.66} &
  \multicolumn{1}{c|}{26.32/2.21} &
  \multicolumn{1}{c|}{17.86/3.63} &
  \multicolumn{1}{c|}{8.54/6.84} &
  \multicolumn{1}{c|}{11.46/0.94} &
  \multicolumn{1}{c|}{7.11/02.03} &
  \multicolumn{1}{c|}{13.67/1.7} &
  {12.46/2.56} \\ \hline
\end{tabular}%
}
\end{table*}

\begin{figure*}[]
\centering
\includegraphics[scale= 0.17]{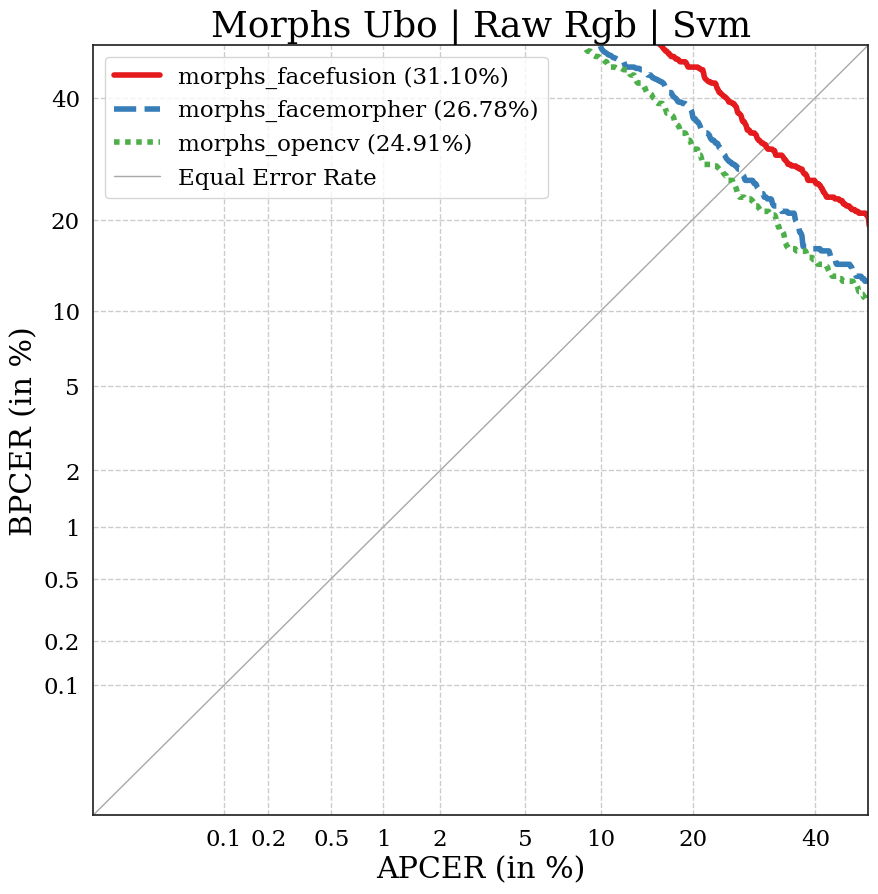}
\includegraphics[scale= 0.17]{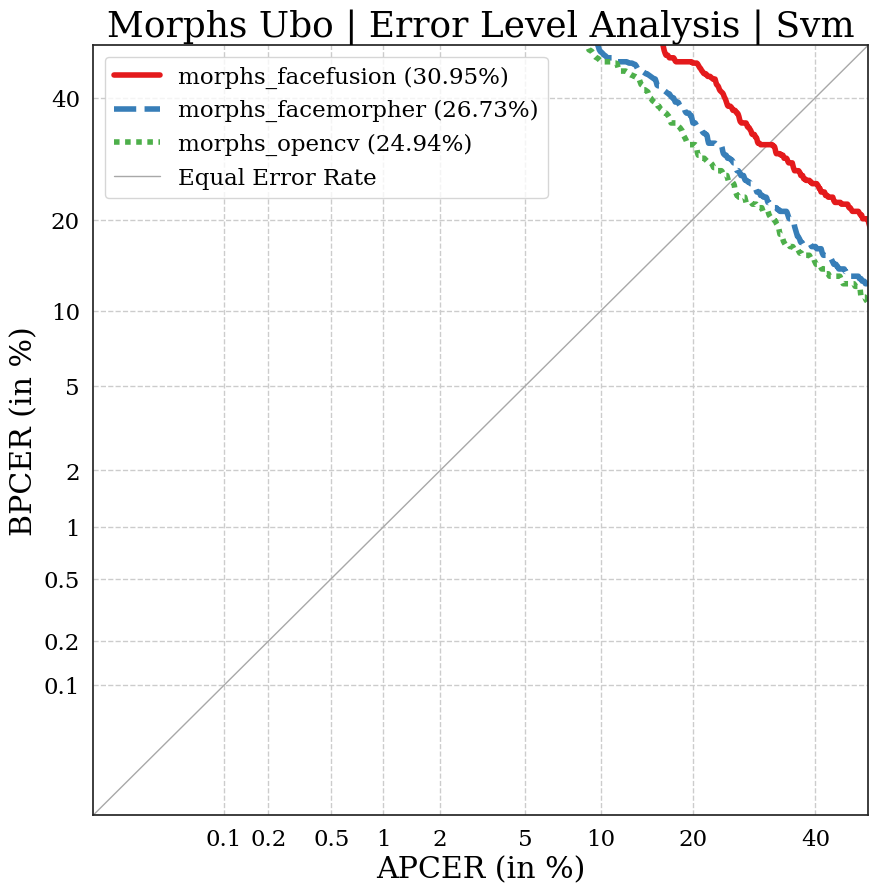}
\includegraphics[scale= 0.17]{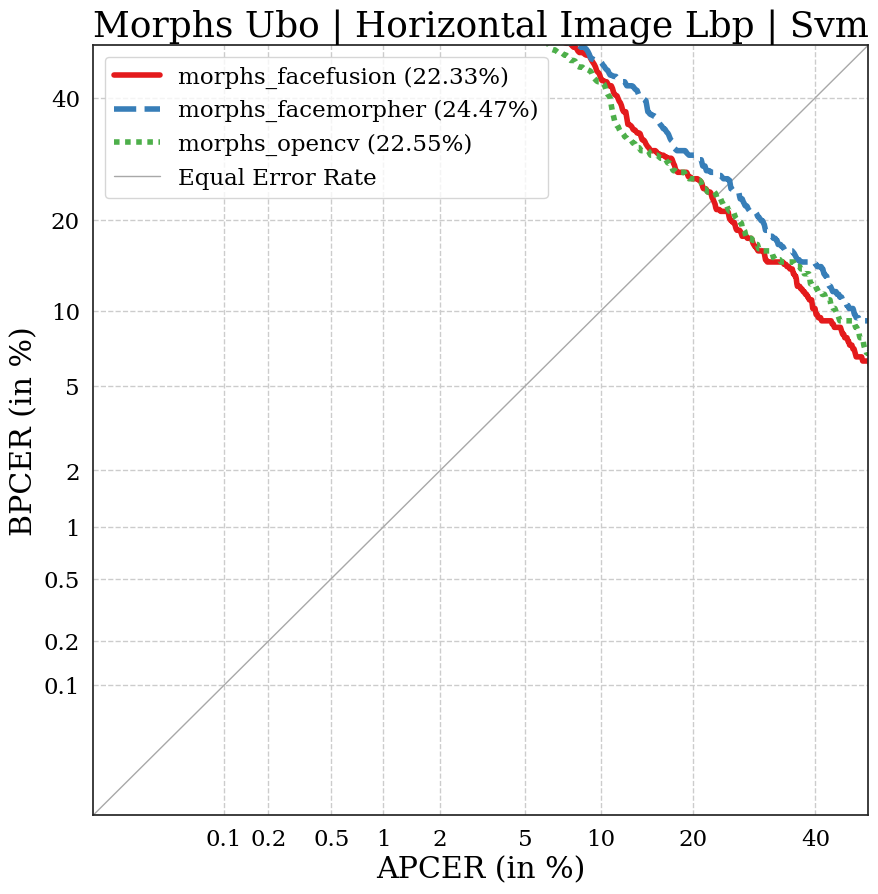}
\includegraphics[scale= 0.17]{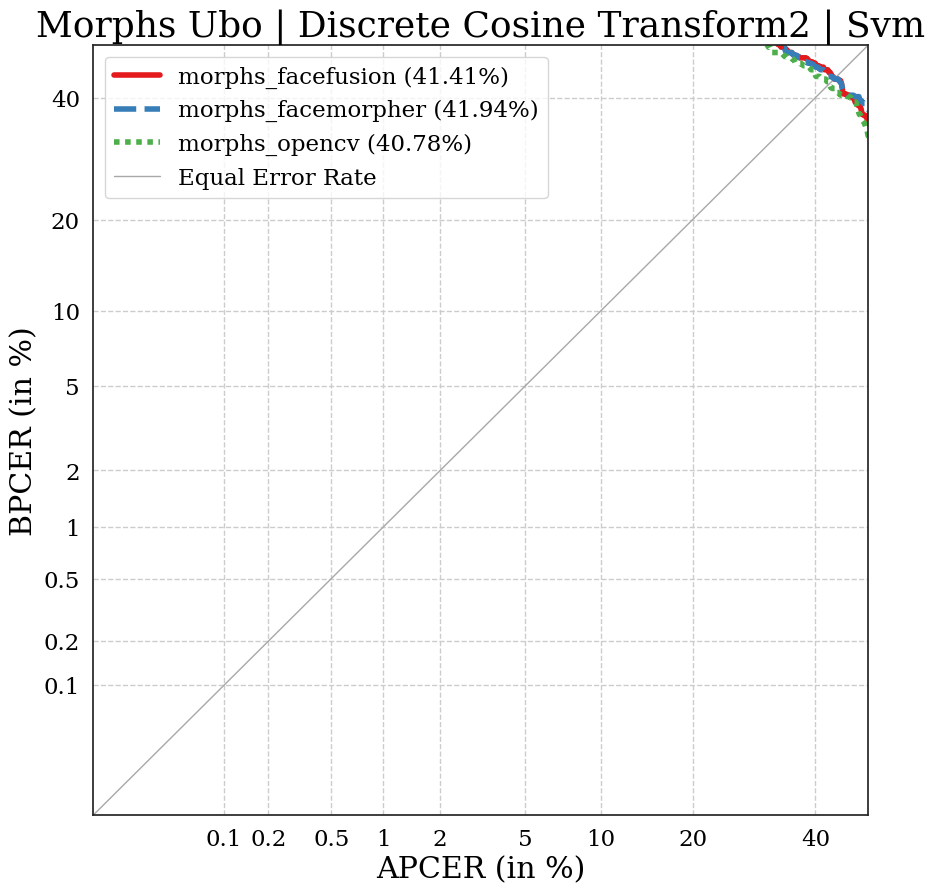}\\

\includegraphics[scale= 0.17]{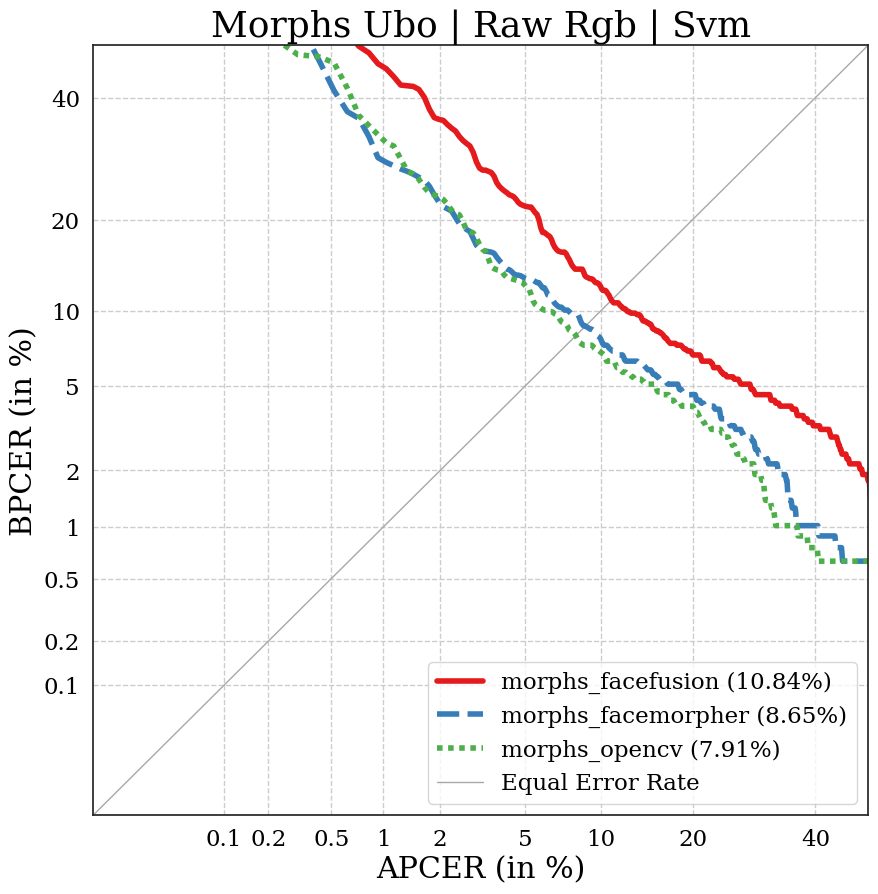}
\includegraphics[scale= 0.17]{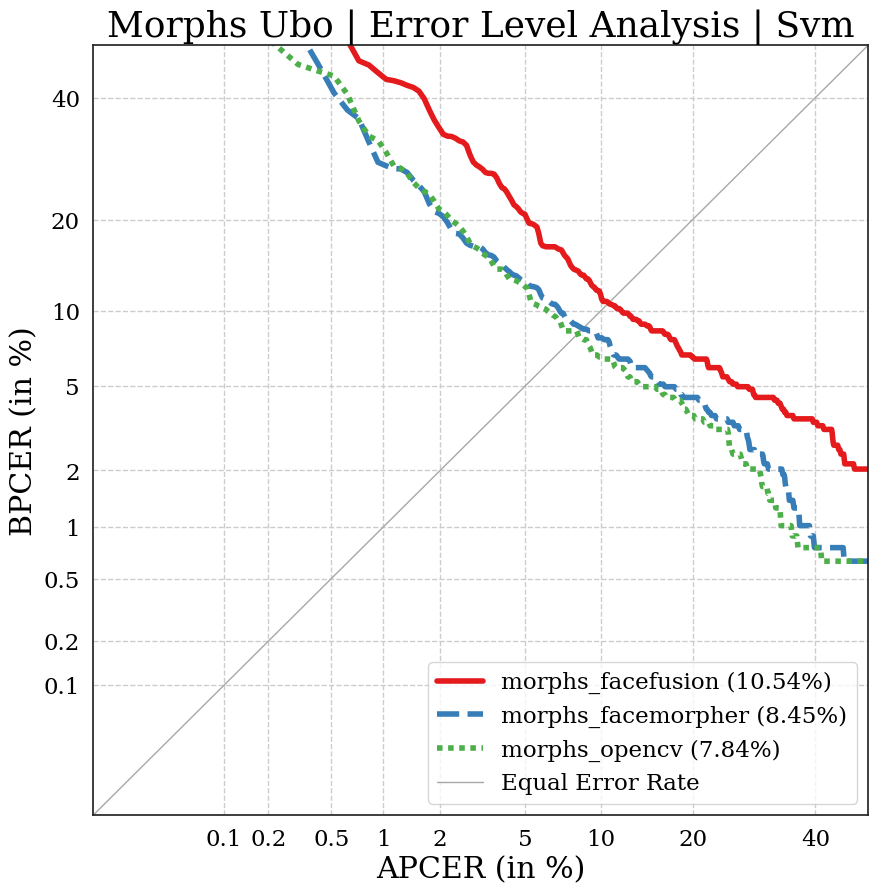}
\includegraphics[scale= 0.17]{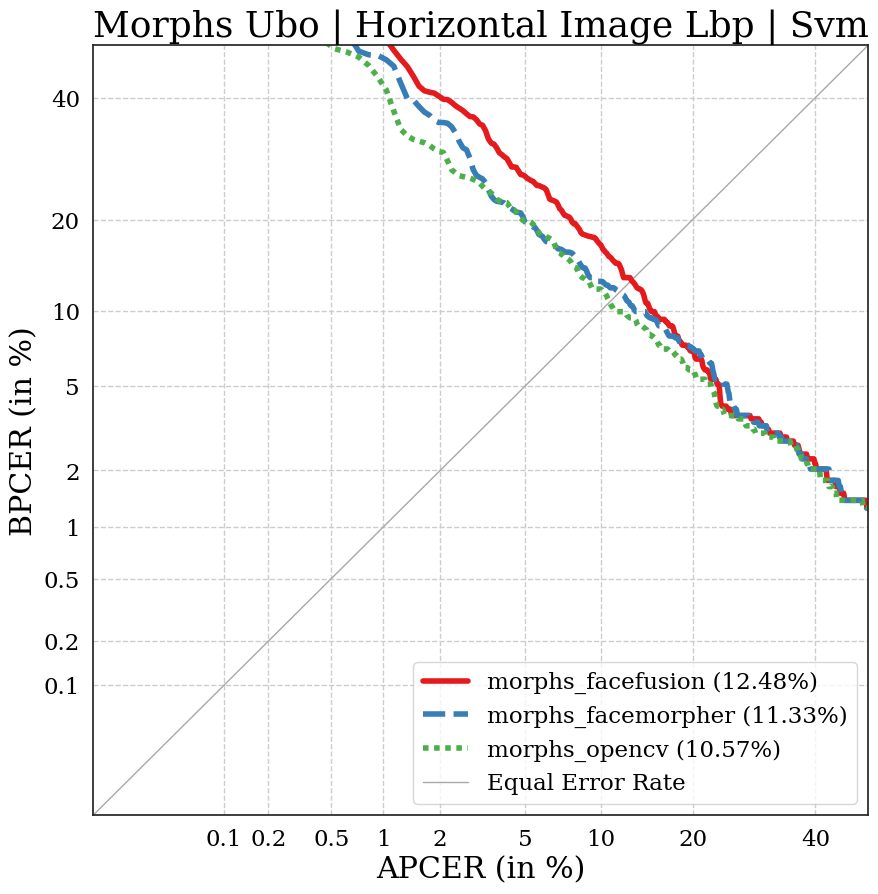}
\includegraphics[scale= 0.17]
{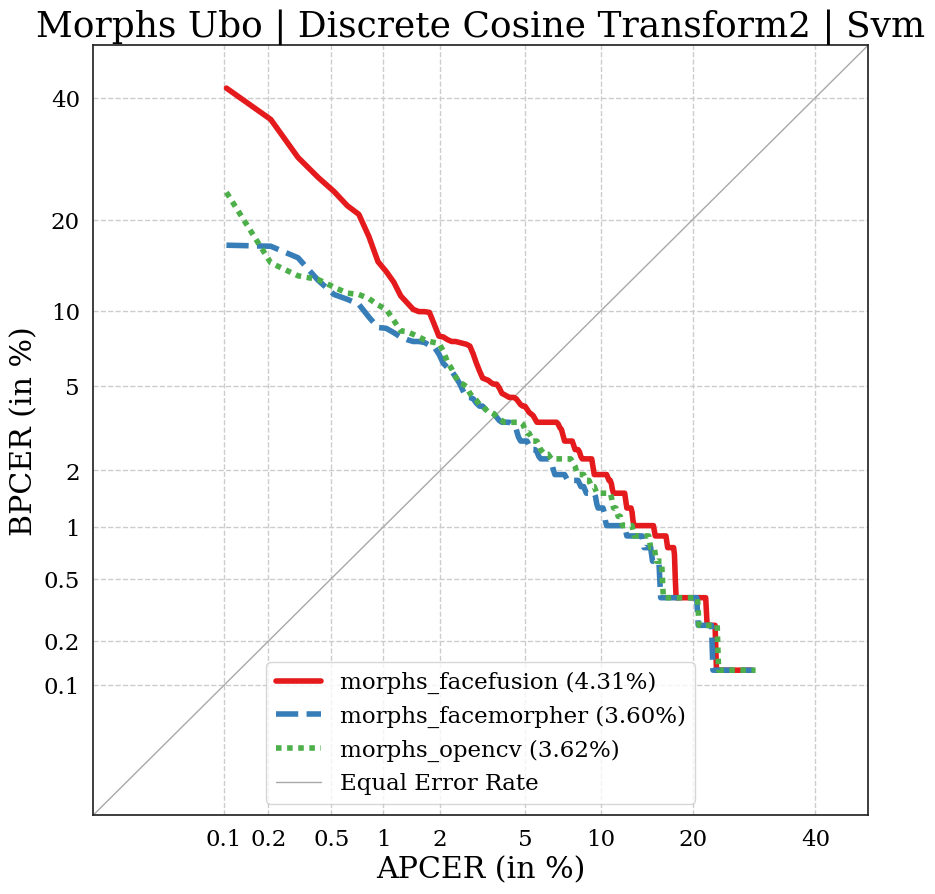}\\

\includegraphics[scale= 0.17]{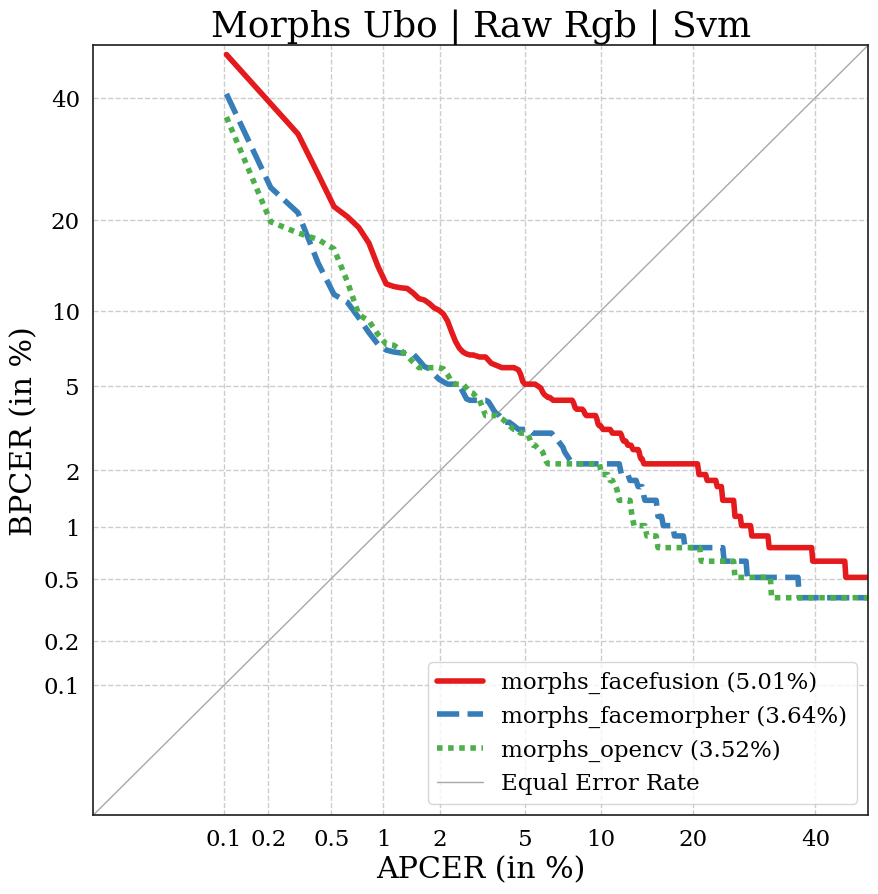}
\includegraphics[scale= 0.17]{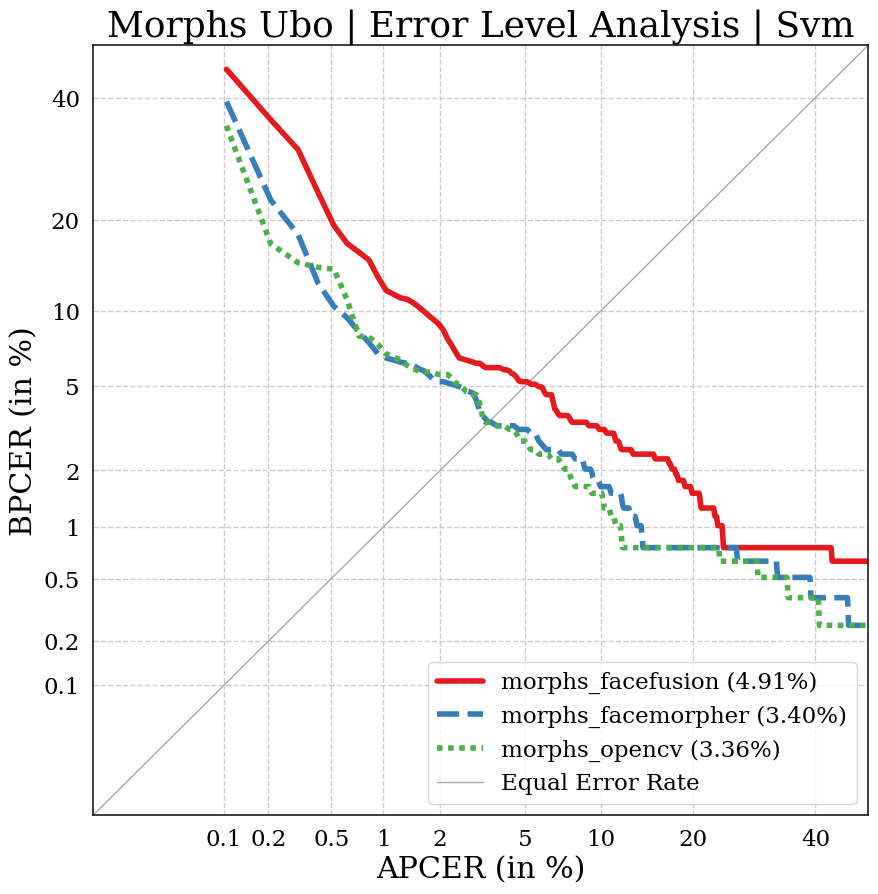}
\includegraphics[scale= 0.17]{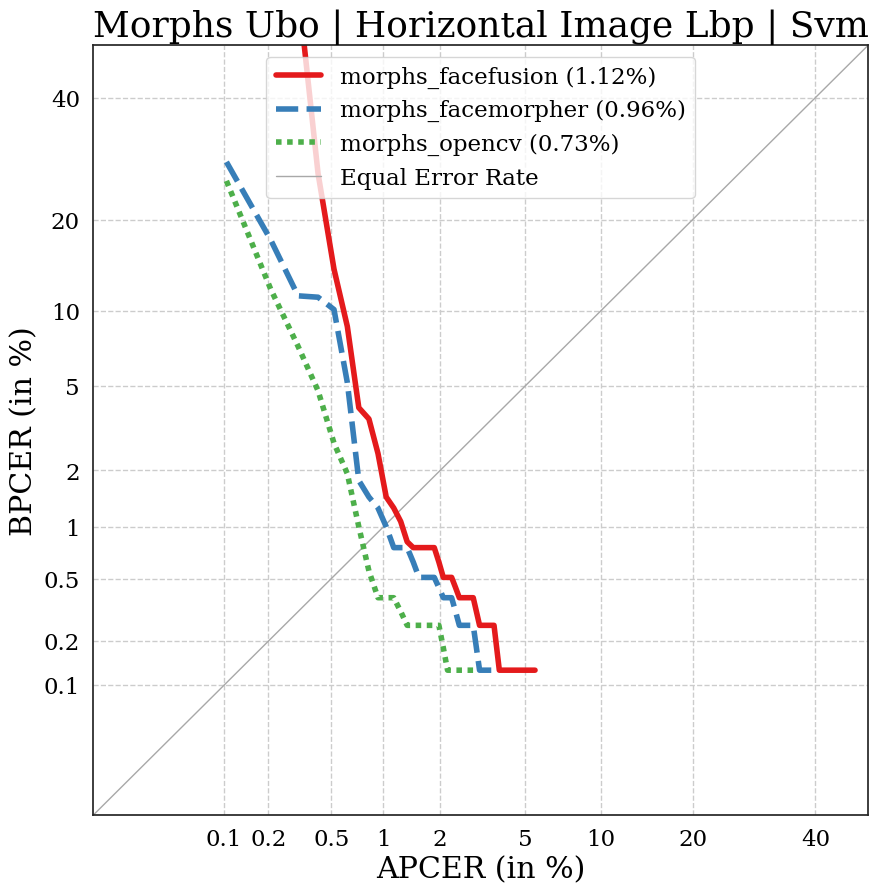}
\includegraphics[scale= 0.17]{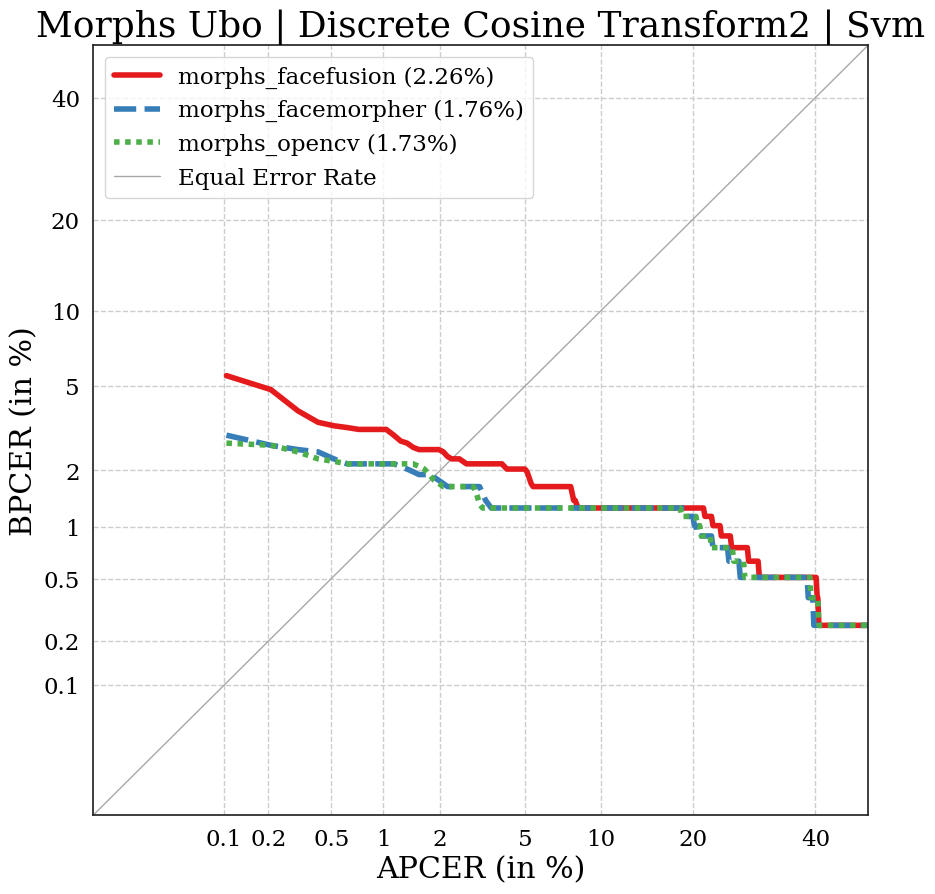}\\

\caption{Top: DET Curves for SVM implementation for four different features extracted (Baseline). Middle: DET for four different features extracted plus Synthetic image generated by CycleGAN (Exp. 2). Bottom: DET Curves for four different features extracted plus texture-transfer technique (Exp. 3).}
\label{fig:det2024}
\end{figure*}

Figure \ref{fig:det2024} shows the DET curves for UBO-Morpher, one of the most challenging in the SOTA in a LOO protocol for SVM with Experiments 2 and 3. The EER for each morphing tool is shown in parentheses. The four DET curves show the MAD results between PS600 handcrafted, PS600 handcrafted plus synthetic (CycleGAN) and PS600 handcrafted plus texture-transfer generated. FaceMorpher and OpenCV were identified as the less challenging morphing tools based on Average results by the "Morphed method", as shown in Table \ref{tab:ML-summary}. Also, the DCT and SRM features extracted were identified as the most suitable for classifying print/scan morphed images according to the Average by "Feature extracted" results applied to all the morphed images. 

\begin{figure*}[]
\centering
\includegraphics[scale=0.18]{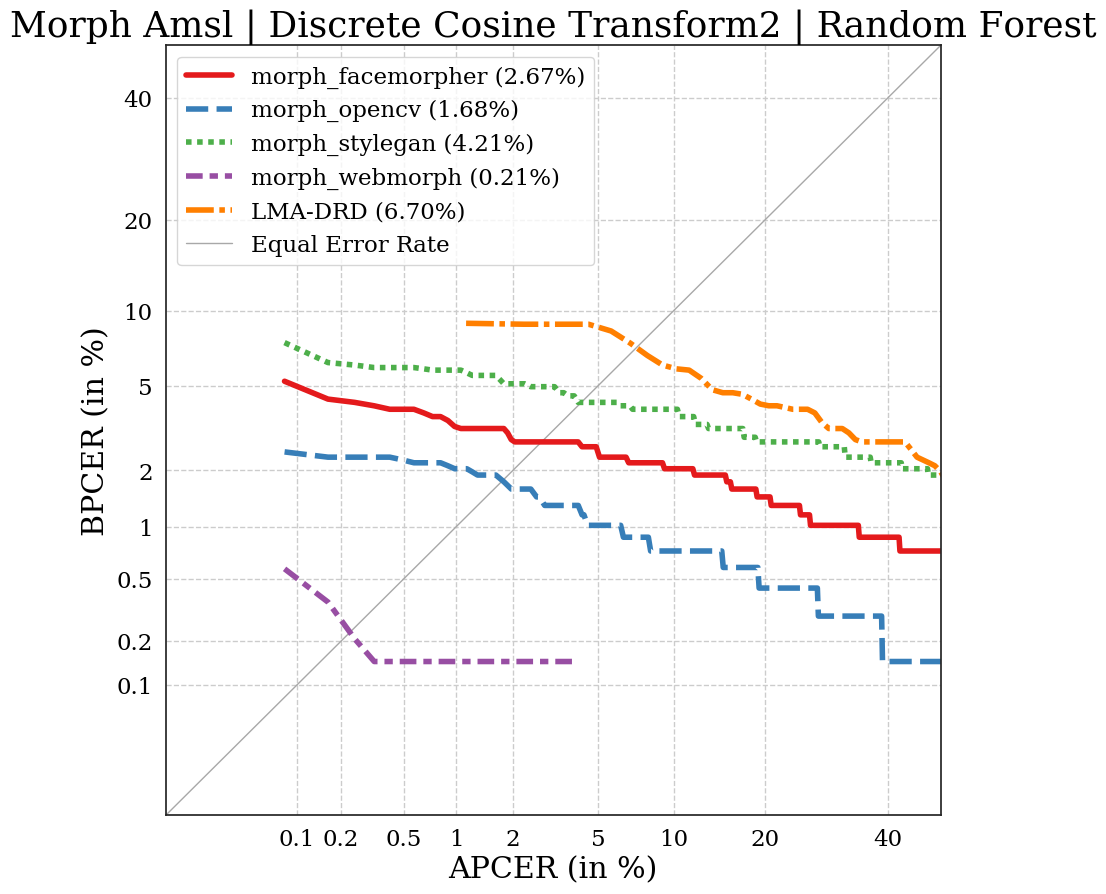}
\includegraphics[scale=0.18]{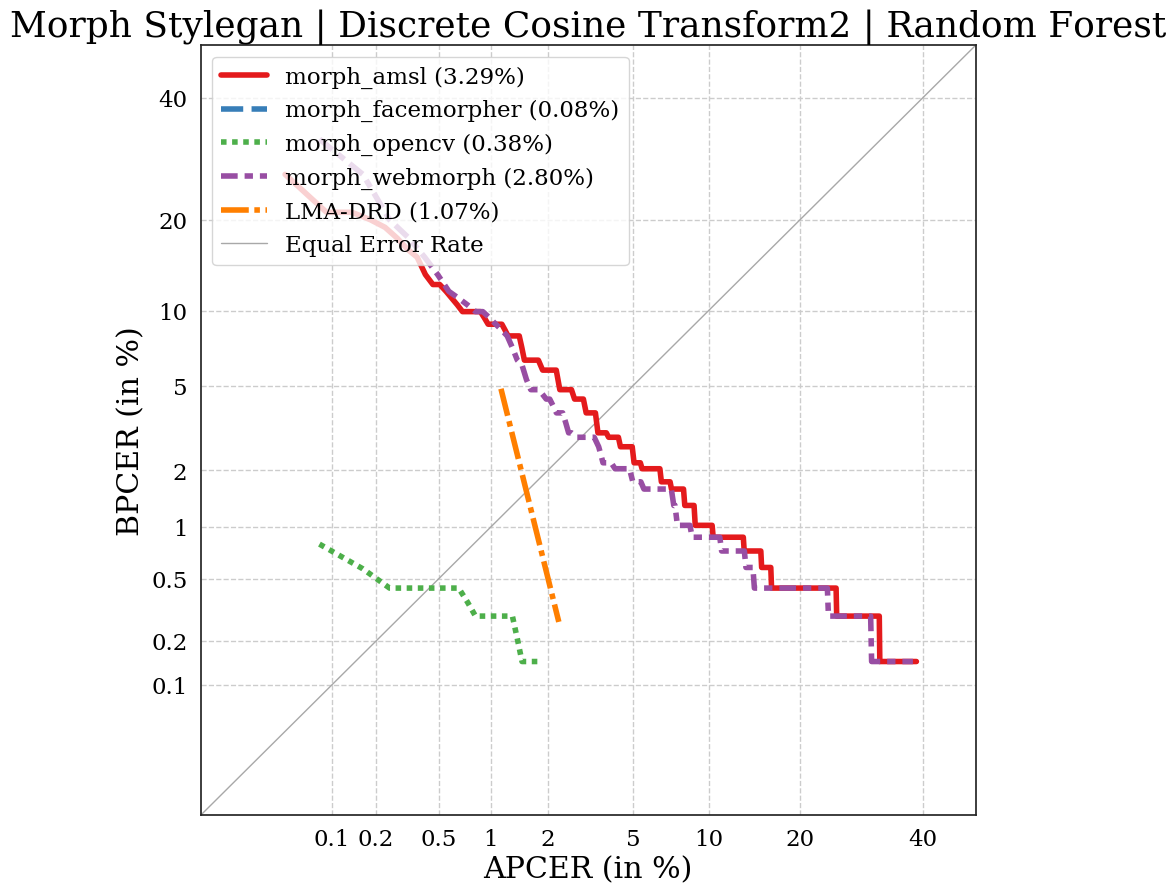}
\includegraphics[scale=0.18]{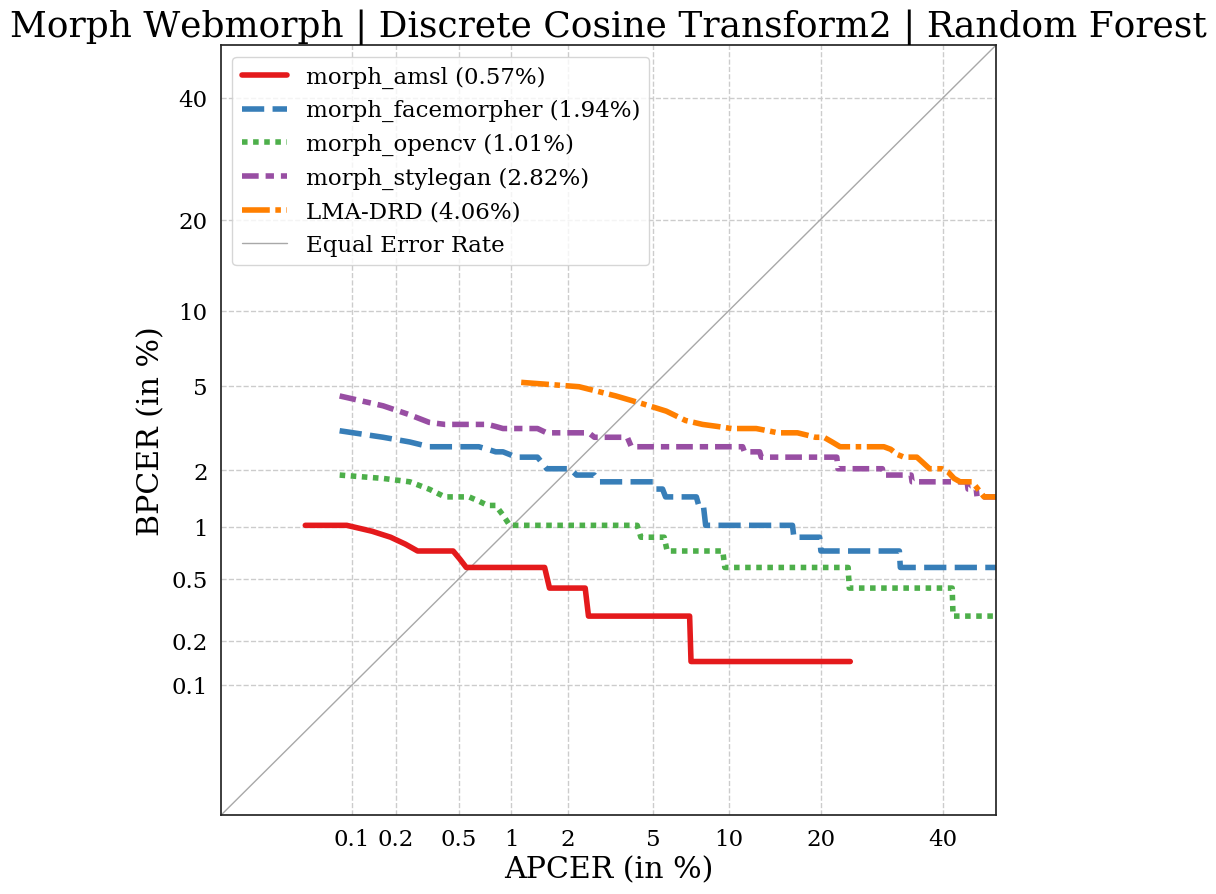}
\caption{Left. DET curve for the AMSL in LOO as a test set using a Random Forest Classifier. Middle: DET curve for the StyleGAN in LOO as a test set using a Random Forest Classifier. Right: DET curve for the WebMorph in LOO as a test set using a Random Forest Classifier.}
\label{fig:cross-All}
\end{figure*}

In order to compare our results with the SOTA, we evaluated the generalisation capability of our approach with FERET/FRGC+CycleGAN+Texture-transfer as a training set and tested in the SOTA open-set datasets in a challenging cross-dataset with FRLL \cite{FRLL}, LMA-DRD and AMSL\cite{AMSL} datasets as a test set. This dataset also contains morph images created by a different morphing tools such as StyleGAN and Webmorph, showing the generalisation capabilities of our model trained in SVM and also in Random Forrest as shown in Figure \ref{fig:cross-All} for AMSL. Also, Figure \ref{fig:cross-All} shows the results for StyleGAN and for Webmorph. The DET curves show the results for the DCT feature extracted. All the evaluations also included the LMA-DRD  test set dataset.

Table \ref{tab:ML-summary} shows the EER for each feature applied to PS600 handcrafted plus Synthetic (Exp. 2) and PS600 handcrafted plus texture-transfer (Exp. 3) on the FRGC/FERET database respectively. Overall, this table shows the best results by "Feature extracted" and the best results by "Morph method" highlighted in bold in a LOO protocol.

According to the results, the classifier trained in frequency components such as DCT, and SRM reached the lowest EER than intensity or shape. On the other hand, the semi-automatically method-based texture can be detected more easily than print/scan images generated by CycleGAN. It is essential to highlight that both methods allow us to improve the results in comparison with the baseline. The comparison can be checked directly from Figure \ref{fig:det2024}.

\vspace{-0.3cm}
\section{Conclusions}
\label{sec:conclusions}

This work shows that creating print/scan from digital databases is feasible to improve the MAD and increase the number of images and scenarios available from training more robust classifiers and developing generalisation capabilities. It is essential to point-out that generated "art images", as proposed in the general applications for GANs, are very different to generated images with real contexts and details such as faces. In this case, any artefact may change the prediction task's results and accuracy. Further on that, the CycleGAN based on ResNet50 and pixel-wise approach allows us to improve even more, reach lower FID scores, and support the S-MAD classifier results in a LOO. On the other hand, Frequency features such as SRM and DCT are identified as more suitable for the detection task. Thus, we can get more realistic images and reduce the EER. The differences between the generated images and baseline reduce the EER by at least 20\%.

As a future work, we will extend this generation process to a larger dataset to train a state-of-the-art CNN method that will be evaluated in the BOEP platform \footnote{\url{https://biolab.csr.unibo.it/fvcongoing/UI/Form/BOEP.aspx}}.


\bibliographystyle{IEEEtran}
\bibliography{References}
\vspace{-0.3cm}

\begin{IEEEbiography}[{\includegraphics[width=0.9in,height=1.25in,clip,keepaspectratio]{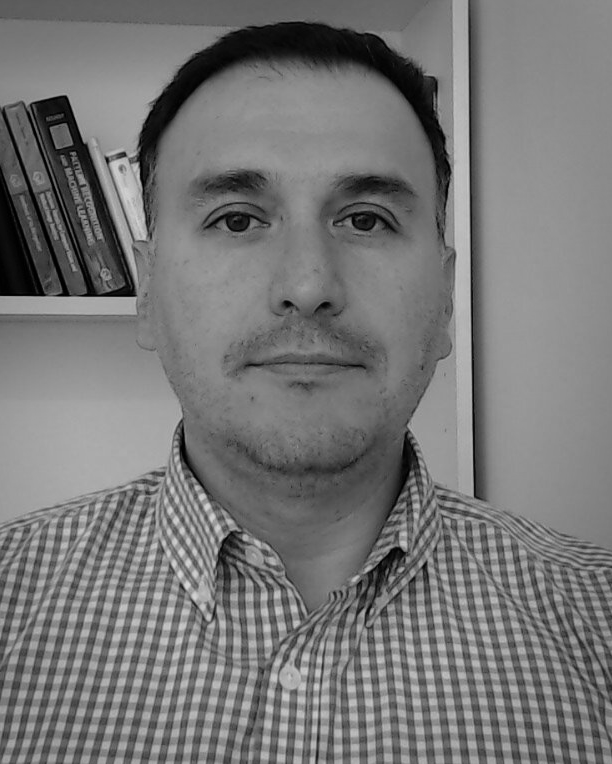}}]{Juan Tapia} received the P.E. degree in electronics engineering from Universidad Mayor, in 2004, and the M.S. and Ph.D. degrees in electrical engineering from the Department of Electrical Engineering, Universidad de Chile, in 2012 and 2016, respectively. In addition, he spent one year of internship with the University of Notre Dame. In 2016, he received the Award for Best Ph.D. Thesis. From 2016 to 2017, he was an Assistant Professor at Universidad Andres Bello. From 2018 to 2020, he was the Research and Development Director for the electricity and electronics area with INACAP, Universidad Tecnologica de Chile, the Research and Development Director of TOC Biometrics Company, and an International Consultor on biometrics for face, iris applications and forensic/tampering ID-card detection. He is currently an Entrepreneur and a Senior Researcher with Hochschule Darmstadt (HDA), leading EU projects, such as iMARS, EINSTEIN and CarMen. His main research interests include pattern recognition and deep learning applied to iris biometrics, morphing, feature fusion, and feature selection. He serves as a reviewer for a number of journals and conferences. He is on behalf of the German DIN Member of the ISO/IEC Sub-Committee 37 on biometrics.
\end{IEEEbiography}
\vspace{-0.3cm}

\begin{IEEEbiography}[{\includegraphics[width=0.9in,height=1.25in,clip,keepaspectratio]{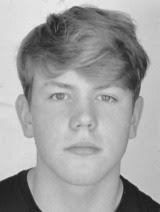}}]{Maximilian Russo} is currently a computer science student at Hochschule Darmstadt (HDA). Alongside his studies, he works as a student assistant with the da/sec research group, which focuses on security and privacy research. He has contributed to various projects, including Android encryption, decryption techniques, and fingerprint scan technology. Currently, he is contributing as a research assistant to explore digital print and scan technologies.
\end{IEEEbiography}
\vspace{-0.3cm}

\begin{IEEEbiography}[{\includegraphics[width=0.9in,height=1.25in,clip,keepaspectratio]{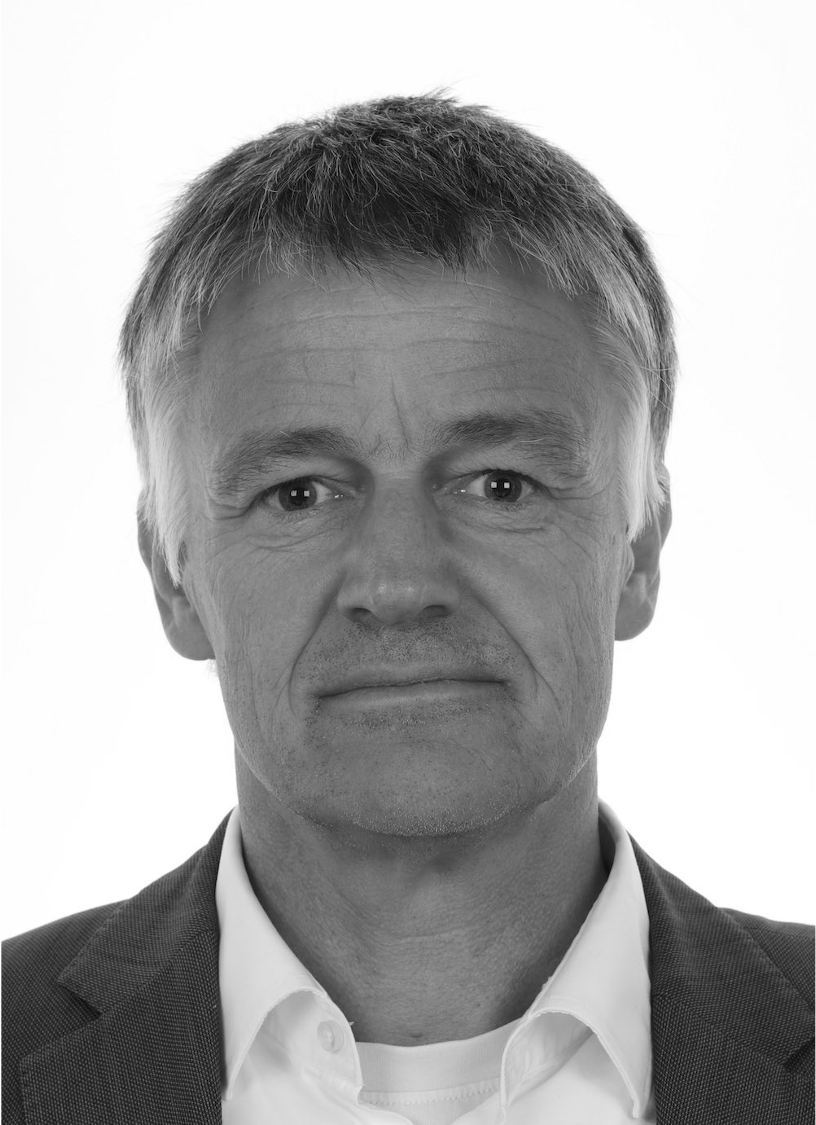}}]{Christoph Busch} is a member of the Department of Information Security and Communication Technology (IIK) at the Norwegian University of Science and Technology (NTNU), Norway. He holds a joint appointment with the computer science faculty at Hochschule Darmstadt (HDA), Germany. Further, he lectures the course Biometric Systems at Denmark’s DTU since 2007. On behalf of the German BSI he has been the coordinator for the project series BioIS, BioFace, BioFinger, BioKeyS Pilot-DB, KBEinweg and NFIQ2.0. In the European research program, he was the initiator of the Integrated Project 3D-Face, FIDELITY and iMARS. Further, he was/is partner in the projects TURBINE, BEST Network, ORIGINS, INGRESS, PIDaaS, SOTAMD, RESPECT and TReSPAsS. He is also principal investigator at the German National Research Center for Applied Cybersecurity (ATHENE). Moreover Christoph Busch is co-founder of the European Association for Biometrics (www.eab.org) which was established in 2011 and assembles in the meantime more than 200 institutional members. Christoph co-authored more than 700 technical papers and has been a speaker at international conferences. He is a member of the editorial board of the IET journal.
\end{IEEEbiography}

\vfill

\end{document}